\documentclass[Afour,sageh,times]{sagej}

\usepackage{moreverb,url}
\usepackage[colorlinks,bookmarksopen,bookmarksnumbered,citecolor=red,urlcolor=red]{hyperref}

\usepackage[caption=false,font=normalsize,labelfont=sf,textfont=sf]{subfig}

\usepackage{empheq}
\usepackage{caption}
\usepackage{import}

\newcommand\BibTeX{{\rmfamily B\kern-.05em \textsc{i\kern-.025em b}\kern-.08em
T\kern-.1667em\lower.7ex\hbox{E}\kern-.125emX}}

\def\volumeyear{2023}
\DeclareMathOperator{\atantwo}{atan2}
\DeclareMathOperator{\acos}{acos}

\begin{document}

\runninghead{Milazzo et al.}

\title{Modeling and Control of a Novel\\ Variable Stiffness Three DoFs Wrist}

\author{Giuseppe Milazzo\affilnum{1}, Manuel Giuseppe Catalano\affilnum{1}, Antonio Bicchi\affilnum{1,2}, and Giorgio Grioli\affilnum{1,2}}

\affiliation{\affilnum{1}Soft Robotics for Human Cooperation and Rehabilitation, Istituto Italiano di Tecnologia, Genova, Italy\\
\affilnum{2}Centro di Ricerca Enrico Piaggio, Università di Pisa, Pisa, Italy.}

\corrauth{Giuseppe Milazzo, Soft Robotics for Human Cooperation and Rehabilitation,
Istituto Italiano di Tecnologia,
Via S. Quirico, 19d,
16163, Genova, Italy.
}

\email{giuseppe.milazzo@iit.it}

\begin{abstract}
This study introduces an innovative design for a Variable Stiffness 3 Degrees of Freedom actuated wrist capable of actively and continuously adjusting its overall stiffness by modulating the active length of non-linear elastic elements. This modulation is akin to human muscular cocontraction and is achieved using only four motors. The mechanical configuration employed results in a compact and lightweight device with anthropomorphic characteristics, making it potentially suitable for applications such as prosthetics and humanoid robotics. This design aims to enhance performance in dynamic tasks, improve task adaptability, and ensure safety during interactions with both people and objects.

The paper details the first hardware implementation of the proposed design, providing insights into the theoretical model, mechanical and electronic components, as well as the control architecture. System performance is assessed using a motion capture system. The results demonstrate that the prototype offers a broad range of motion ($[55, -45]$° for flexion/extension, $\pm48$° for radial/ulnar deviation, and $\pm180$° for pronation/supination) while having the capability to triple its stiffness. Furthermore, following proper calibration, the wrist posture can be reconstructed through multivariate linear regression using rotational encoders and the forward kinematic model. This reconstruction achieves an average Root Mean Square Error of {6.6°}, with an $R^2$ value of {0.93}.
\end{abstract}


\maketitle

\section{Introduction}
One of the challenges of modern robotics is to craft machines that can physically interact and cooperate with people.
\begin{figure}[!t]
    \centering
          \includegraphics[trim=20cm 0cm 10cm 0cm,clip, width =0.93\linewidth]{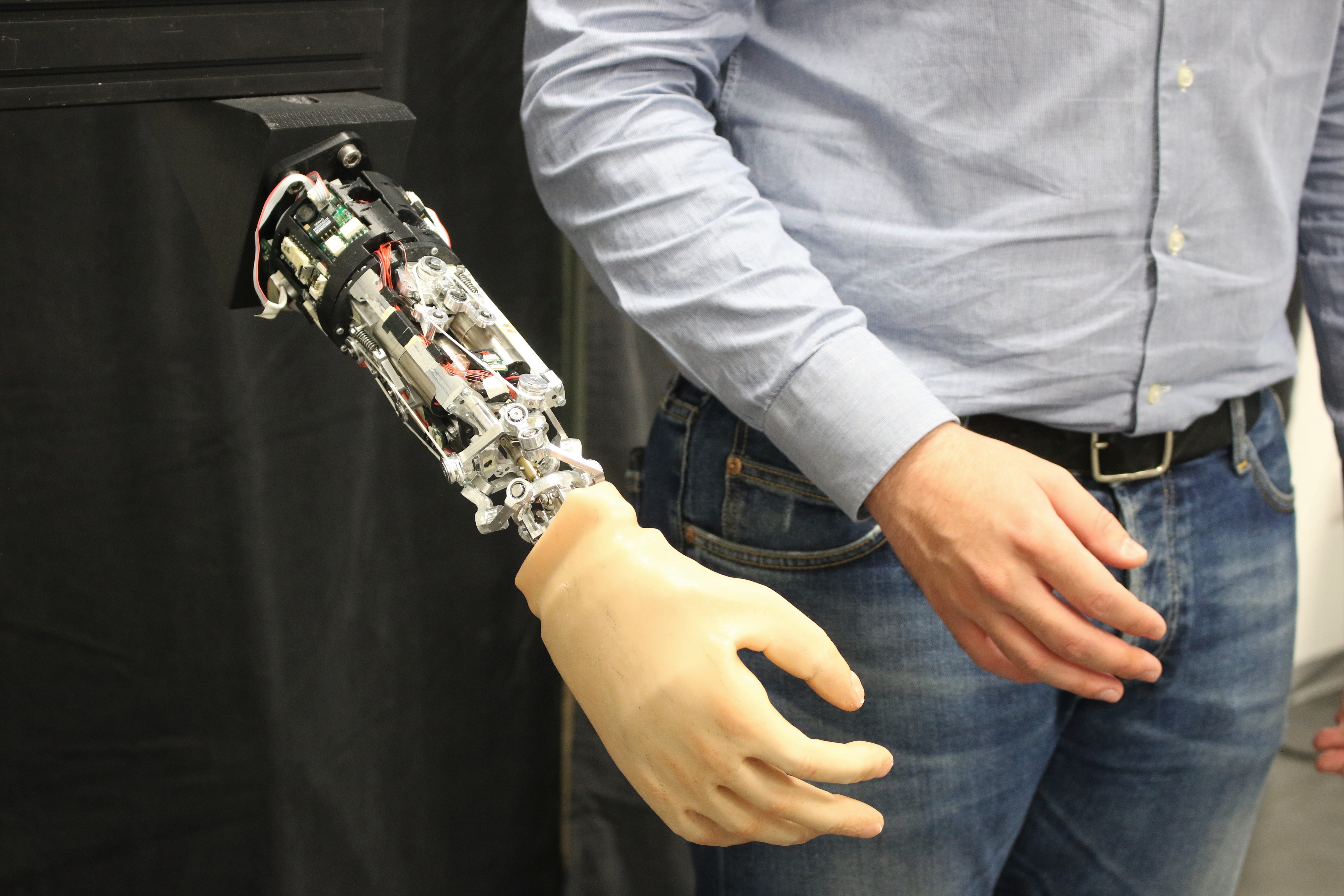}
    \caption{The VS-Wrist compared to a human forearm (model’s size: forearm length 247 mm, forearm circumference 265 mm. Comprised between the 5th and 50th percentile male \cite{nasa_dim}).}
    \label{fig:VS-Wrist}
\end{figure}
Taking inspiration from humans, compliant end-effectors have proven their capability to adapt using simple actuation and control systems in various applications, ranging from industrial \cite{firth2022anthropomorphic}, \cite{zongxing2020research}, to prosthetics \cite{tavakoli2014adaptive} and service robotics \cite{trobinger2021introducing}. 
However, to promote the simplicity of the hardware, robotic limbs usually equip rigid and underactuated serial wrist joints \cite{soawrists}, \cite{fan2022prosthetic}. Nonetheless, wrist articulation plays a crucial role in manipulation tasks since it allows for varying the hand pose to reach the best posture for grasping an object.
Therefore, a complete wrist joint should {have} 3 DoFs to freely orient the end-effector in a tridimensional space. 
\newline \indent Many studies prove that human beings adjust the stiffness of their limbs by exploiting muscular cocontraction to adapt to tasks of different natures. In summary, a large stiffness performs better in resisting perturbations and accomplishing precision tasks, while a soft behavior is more suitable for those assignments that require exploring an unknown environment with small interaction forces \cite{blank2014task}, \cite{borzelli2018muscle}, \cite{osu2004optimal}. 
Therefore, it is desirable to replicate this feature of controllable impedance also in robots. One possible way is to adjust the stiffness of rigid manipulators via software, taking advantage of established impedance control techniques \cite{Hogan1985ImpedanceCA}, \cite{yellewa2022design}. 
\begin{figure*}[!t]
    \centering
    \subfloat[]{\label{fig:General_Wrist_Scheme}\includegraphics[height = 16em]{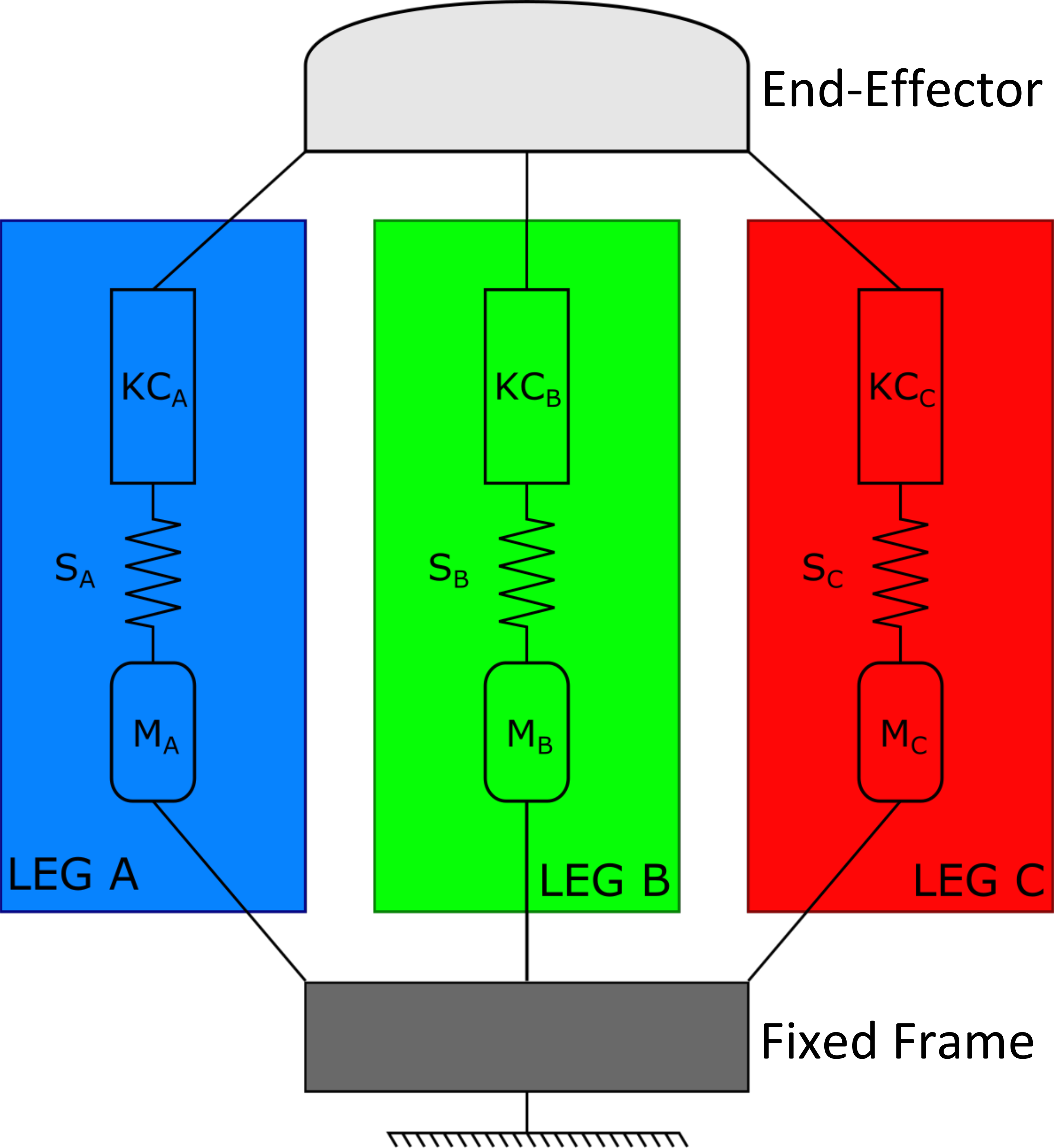}}
    \hfil
    \subfloat[]{\label{fig:Wrist_dof}\includegraphics[height = 16em]{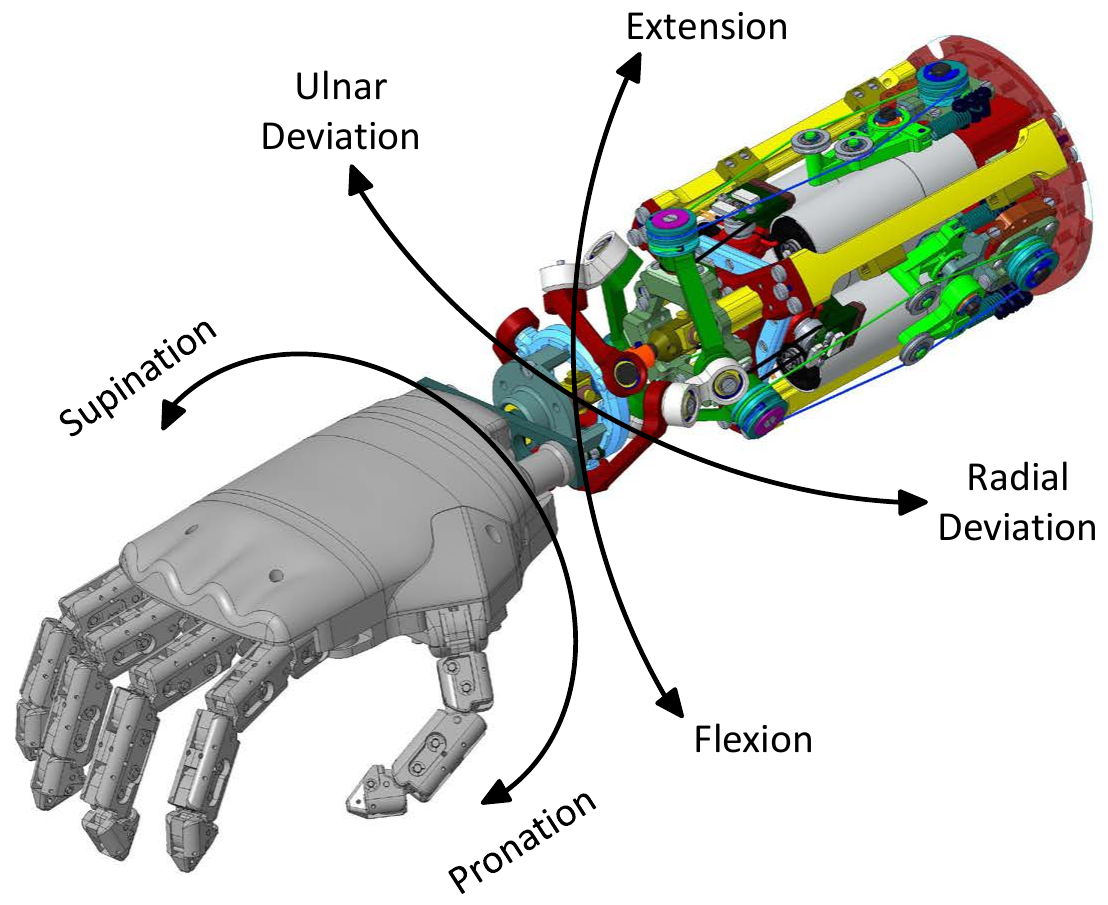}}
    \hfil
    \subfloat[]{\label{fig:vs_W}\includegraphics[height = 16em]{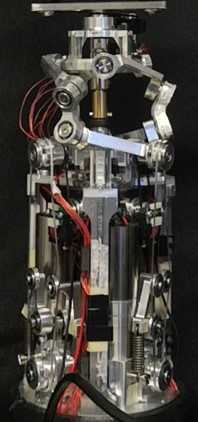}}
    \caption{Panel (a) depicts a schematic architecture of the 2 DoFs VS joint integrated into the VS-Wrist. Its kinematic configuration is a parallel manipulator that achieves hemispherical movements of the coupler. Each leg, distinguished by a specific color, consists of a motor unit, a non-linear elastic transmission, and a kinematic chain of four revolute joints. Panel (b) displays the CAD of the VS-Wrist equipping a robotic prosthetic hand and highlights its DoFs. Panel (c) shows a close-up picture of the prototype.}
    \label{fig:General_Wrist}
\end{figure*}

Another possibility {involves} employing soft robots that embed Variable Stiffness (VS) in {their} hardware implementation. In contrast to classical machines, soft robots are {better suited} for applications {involving interactions} with the environment, {prioritizing safety and robustness over the power and precision requirements typical of industrial manipulators.}
Articulated soft robots achieve this goal by embedding compliant elements in their actuation or transmission, resulting in a behavior comparable to the musculoskeletal system of vertebrates \cite{migliore2005biologically}. 
In addition, using redundant non-backdrivable actuators, the elastic elements can apply constant forces to increase the stiffness of the joint without requiring a continuous input of energy. 

Passive stiffness regulation {offers} several advantages when compared to impedance control. Active systems use {joint movement and interaction forces measurements} to modulate feedback gains, changing the effective joint stiffness. 
However, actively controlling impedance in torque-controlled robots and backdrivable variable stiffness joints requires either high computational cost or a constant energy drain. {These} drawbacks are particularly relevant to mobile robots and prostheses, where the entire device must match restrictive size and weight constraints that limit the computational power and battery capacity \cite{ENGLISH19997}.
Moreover, \cite{haddadin2007safety} shows that, concerning safety, even if impacts are detected timely, the motors could not react {quickly enough with} solely an impedance control. {Therefore,} the system should be considered stiff during impacts, making impedance-controlled robots potentially dangerous.
{In contrast}, compliant mechanisms protect the device {during} external impacts by decoupling the output link from the rest of the system. Furthermore, soft robots are more robust since they can remain compliant even when the actuators are disabled or faulty \cite{SoftRobotics}. 
{Additionally}, the energy stored in the elastic elements could reduce power consumption and enhance motor abilities in explosive movements and cyclic tasks \cite{SoftRobotics}, \cite{HADDADIN20116863}, \cite{grioli2015variable}. \\
\indent Despite the numerous advantages of VS actuation,  artificial wrists commonly lack this feature. Some commercial passive wrists\endnote{Össur,
i-Limb Quantum Flexion Wrist, \url{https://irp-cdn.multiscreensite.com/acf635e2/files/uploaded/Flexion\%20Wrist.pdf}. Last Accessed: May 2023.} embed elastic elements to comply along flexion/extension movements \cite{archer2011wrist}. Yang et al. present a serial 2 DoFs soft wrist made of the sequence of two continuum bending and torsion modules, which can vary the stiffness of the bending joint by inflating a balloon \cite{soft_VS}.
Von Drigalski et al. show a passive 6 DoFs wrist for industrial applications with inherent compliance that can switch to the rigid configuration using an external actuator \cite{softwrist_6dof}. Recently, researchers employed powered one DoF VS joints for elbow \cite{vs_elbow}, \cite{Baggetta2022} and ankle prostheses \cite{VS_powered_ankle}, and motor rehabilitation devices \cite{VS_ExoRehab}, \cite{knee_rehab_VS}, exploiting their affinity with the human musculoskeletal system.
However, there still is a lack of fully actuated 3 DoFs wrist joints with variable stiffness. 

This work presents a novel and compact VS wrist with 3 DoFs (shown in Figure~\ref{fig:VS-Wrist}) that can actively vary its overall stiffness with continuity. Thanks to its anthropomorphous characteristics, it holds potential applications in prosthetics and humanoid robotics. This paper extends the mathematical model of a concept already introduced in \cite{lemerle2021wrist} and \cite{lemerle2021soft}, validating the efficacy of the idea through the first hardware implementation.
Section~\ref{sect:design} explains the general design concept of the device, and Section~\ref{sect:model} describes its model. 
Section~\ref{sect:elastic_tr} focuses on the elastic transmission mechanism, describing its working principle and mathematical model. A detailed report of the hardware implementation follows in Section~\ref{sect:hardware}. Section~\ref{sect:control} depicts the employed control strategy, and Section~\ref{sect:calibration} details the system calibration.
Section~\ref{sect:experimental_validation} reports the experimental performance evaluation with a Motion Capture system and showcases practical applications of the device. Finally, Section~\ref{sect:discuss} discusses the experimental results and the applicability of the presented architecture in prosthetics, while conclusions are drawn in Section~\ref{sect:conclusion}.

\section{Design Concept} \label{sect:design}
The VS-Wrist features a hybrid serial-parallel architecture {leveraging} the redundancy of the actuation system and a non-linear elastic transmission to adjust the stiffness of the coupler. It comprises a 2 DoFs VS joint achieved {through} a Parallel Manipulator (PM) architecture {complemented by an independent serial motor unit that is decoupled from the variable stiffness mechanism.}

Figure~\ref{fig:General_Wrist_Scheme} shows the {overall} architecture of the 2 DoFs VS joint. As discussed in \cite{lemerle2021wrist}, its leg kinematics {draws inspiration from} the Omni-Wrist III {outlined} in \cite{Rosheim_OmniWrist}. However, the proposed design {incorporates} only three legs instead of four, and introduces substantial differences in the actuation system to enable the modulation of joint stiffness.
Precisely, the 2 DoFs VS joint {integrates} a supplementary actuator and a non-linear elastic mechanism {in its design} to transmit the motion of each motor to the first joint of each kinematic chain.

Figure~\ref{fig:Wrist_dof} showcases the mechanical design of the VS-Wrist and delineates its DoFs. In this regard, the PM  manages the radial/ulnar deviation (RUD) and the flexion/extension (FE) of the wrist, while the serial motor unit actuates the device along the pronation/supination (PS) axis. Unlike the human wrist, this {latter} motor unit only rotates the hand, {rather than} the entire forearm. 

The VS-Wrist {boasts} an innovative mechanical structure that embodies part of the system intelligence. {Typically}, an $n$ DoFs VS joint would {necessitate} $2n$ motors to control its position and stiffness. {By utilizing a PM, it becomes possible to modulate joint stiffness using internal torques, requiring only three motors to control two positional DoFs and leaving one DoF dedicated to overall stiffness regulation. Consequently}, the PM architecture achieves a two DoFs VS joint with the minimum possible number of motors, {resulting in} a compact and lightweight device. This design choice aligns with the human mechanism of muscular impedance regulation, where muscular cocontraction {predominantly} modulates the compliance ellipsoid dimensions, while the orientation of its axes essentially depends on posture \cite{ajoudani2018reduced}, \cite{lemerle2021wrist}. 

\section{System Analysis}\label{sect:model}
{To investigate the behavior of the device, we commence the kinematic analysis by} considering the PM alone. {Subsequently}, we incorporate the effect of the serial motor unit, {given that} its rotation does not impact the PM kinematics. 
We define the reference coordinate frame $\{S_b\} = \{O_b, X_b, Y_b, Z_b\}$, fixed to the base of the wrist, and $\{S_e\} = \{O_e, X_e, Y_e, Z_e\}$, attached to the coupler. As shown in Figure \ref{fig:UJ_DH}, their origins ($O_b$ and $O_e$) are {located} at the center of the base and coupler, respectively. {The $X$-axes point upwards, the $Z$-axes point to the right, and the $Y$-axes are determined by the right-hand rule}.
The pose of the end-effector with respect to {(w.r.t.)} the fixed frame $\{S_b\}$ can be parametrized by the vector $\textbf{x} = [ \alpha_x \enspace \alpha_y \enspace \alpha_z \enspace x_e \enspace  y_e \enspace z_e]^{\top}$, whose first three components represent Euler angles, while the last three defines the position of $O_e$. However, since the PM has 2 DoFs, the minimum parametrization $u~=~[ \alpha_y \enspace \alpha_z]^{\top}$, {representing} the RUD and the FE angles of the wrist, is sufficient to uniquely determine its pose.

We define the homogeneous transformation matrices $\text{T}_{T_\ast}(p)$ and $\text{T}_{R_\ast}(p)$ that encode translation and rotation along the generic *-axis of the quantity $p$.
{Consistent} with the analysis provided in \cite{lemerle2021wrist}, the matrix $\text{T}_b^e(\textbf{x})$, {encoding} the pose \textbf{x} of $\{S_e\}$ w.r.t. $\{S_b\}$, is
\begin{equation}
    \label{eq:pose_euler_extend}
    \text{T}_b^e (\textbf{x}) =
    \begin{array}{cc}
                  &\text{T}_{T_z}(z_e) \text{T}_{T_y}(y_e) \text{T}_{T_x}(x_e)  \\
                    & \quad   \text{T}_{R_z}(\alpha_z) \text{T}_{R_y}(\alpha_y) \text{T}_{R_x}(\alpha_x) 
    \end{array} \enspace . \\
\end{equation}

\begin{figure}[!t]
    \centering
    \includegraphics[width = \linewidth]{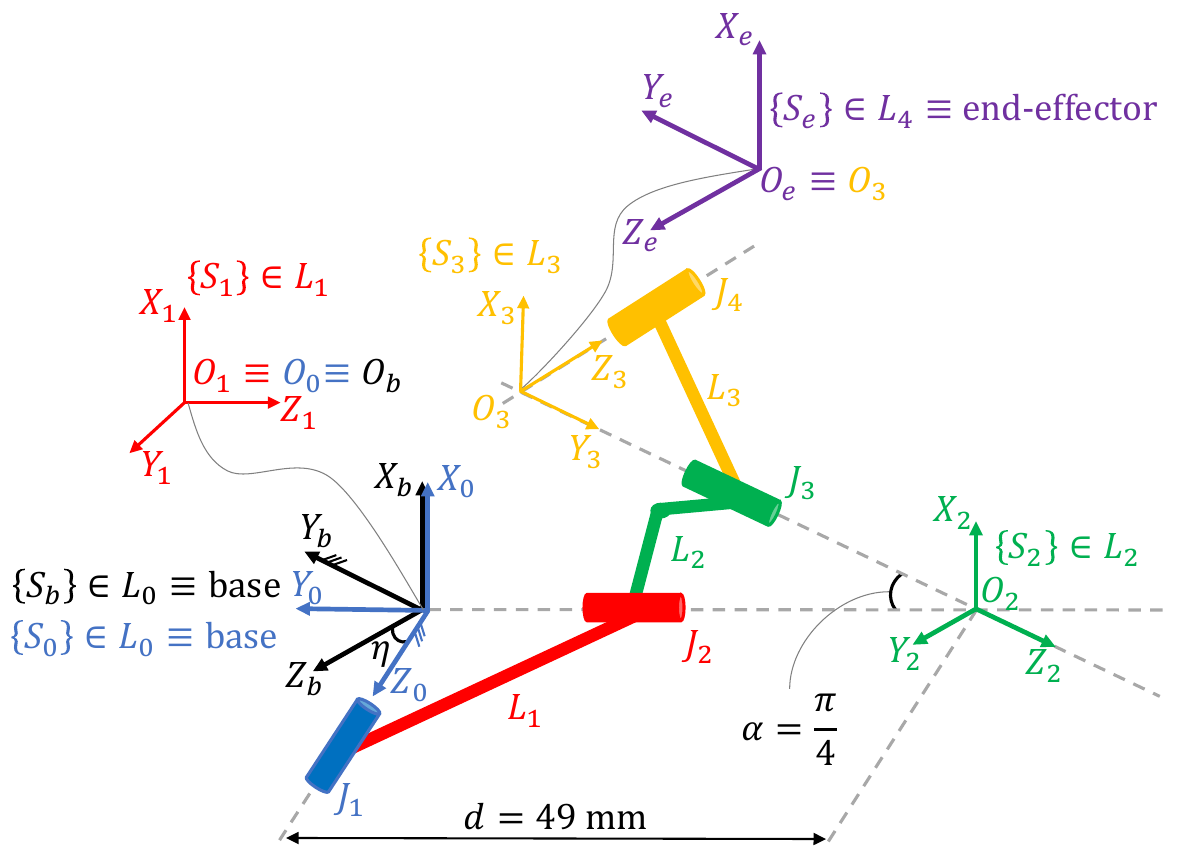}
    \caption{Definition of the local reference coordinate frames of a generic leg according to the Denavit Hartenberg convention. d, $\eta$, and $\alpha$ are fixed design parameters.}
    \label{fig:DH_wrist}
\end{figure}

The transformation matrix $\text{T}_b^e$ can also be {derived} from the joint angles of a single leg $q = [q_1 \enspace q_2 \enspace q_3 \enspace q_4]^{\top} $.
We parametrize {each} kinematic chain of the manipulator according to the Denavit Hartenberg convention, as shown for a generic leg in Figure~\ref{fig:DH_wrist} and Table \ref{tab:DHwrist}.
{The intermediate transformations in local frames are assembled to obtain}
\begin{equation}
    \text{T}_b^e (q) = \text{T}_b^0(\eta) \text{T}_0^1(q_1) \text{T}_1^2(q_2) \text{T}_2^3(q_3) \text{T}_3^e(q_4) \enspace ,
\end{equation}
where $\text{T}_{i-1}^i = \text{T}_{R_z}(\theta_i) \text{T}_{T_z}(d_i) \text{T}_{R_x}(\alpha_i) \text{T}_{T_x}(a_i)$
defines the homogeneous transformation matrix between subsequent local frames.
The PS rotation axis is always parallel to the $X_e$-axis. Therefore, indicating with $\theta_{PS}$ the related motor angle, the resulting homogeneous transformation matrix becomes
\begin{equation}
    \text{T}_b^{e'}(q,\theta_{PS}) = \text{T}_b^e(q) \text{T}_{R_x}(\theta_{PS}) \enspace.
\end{equation}
\begin{table}
\small\sf\centering
\caption{Denavit Hartenberg parametrization of a generic leg of the PM. $d = 49$ mm, $\alpha = \frac{\pi}{4}$ are fixed design parameters.}\label{tab:DHwrist}
\begin{tabular*}{\linewidth}{@{\extracolsep{\fill}} c | c c c c}
\toprule
 $\mathbf{\{S_{i-1}\} \rightarrow \{S_{i}\}}$ & $\mathbf{\theta_i}$ & $\mathbf{d_i}$ & $\mathbf{a_i}$ & $\mathbf{\alpha_i}$\\
\midrule
$\{S_{b}\} \rightarrow \{S_{0}\}$ & 0 & 0 &  0 & $\eta$\\
$\{S_{0}\} \rightarrow \{S_{1}\}$ & $q_1$ & 0 & 0 & ${\pi}/{2}$\\
$\{S_{1}\} \rightarrow \{S_{2}\}$ & $q_2$ & $d$ & 0 & $-\alpha$\\
$\{S_{2}\} \rightarrow \{S_{3}\}$ & $q_3$ & $-d$ & 0 & ${\pi}/{2}$\\
$\{S_{3}\} \rightarrow \{S_{e}\}$ & $q_4$ & 0 &  0 & $\pi - \eta$\\
\bottomrule
\end{tabular*}
\end{table}
\subsection{Pronation/Supination Transmission Kinematics}
\begin{figure}
\centering
\includegraphics[width = 0.95\linewidth]{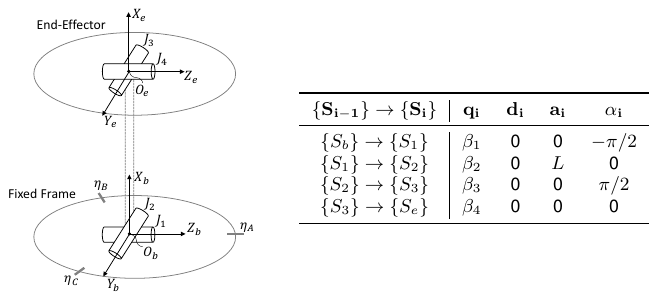}
\caption{Schematic representation of the PS transmission. Each UJ is represented with two perpendicular and incident revolute joints, positioned at the center of either the fixed base frame or the coupler. Note that the two UJs are 90° out of phase to counteract the changing angular velocity of the driving shaft introduced by the lower UJ.
The table on the right summarizes the Denavit Hartenberg parametrization of the kinematic chain. $L = d \sqrt{2(1-\cos(\alpha)} = 37.5$ mm is a fixed design parameter.}
\label{fig:UJ_DH}
\end{figure}

{The PS motor transmits motion to the end-effector through a kinematic chain featuring two opposed Universal Joints (UJs) with rotation centers located at $O_b$ and $O_e$.}
Figure~\ref{fig:UJ_DH} illustrates the schematization of the PS kinematic chain and reports its Denavit Hartenberg parametrization.
Therefore, the UJs angles $\beta= [\beta_1 \enspace \beta_2 \enspace \beta_3 \enspace \beta_4]^{\top}$ parametrize the homogenous transformation matrix from $\{S_b\}$ to $\{S_e\}$ as
\begin{equation} \label{eq:Forward_cardano}
    \text{T}_b^e(\beta) = \text{T}_b^1(\beta_1) \text{T}_1^2(\beta_2) \text{T}_2^3(\beta_3) \text{T}_3^e(\beta_4) \enspace .
\end{equation} 

To solve the Inverse Kinematics (IK) of the PS mechanism, we exploit that the position of the coupler is independent of $\beta_3$ and $\beta_4$, and thus
\begin{equation} \label{eq:UJ_1IK}
    O_2(\beta_1, \beta_2) = \text{T}_b^2(1:3,4) = \text{T}_b^e(1:3,4) = O_e \enspace .
\end{equation}
Solving \eqref{eq:UJ_1IK} yields $\beta_1$ and $\beta_2$ as
\begin{equation}\label{eq:beta12}
\begin{cases}
    \beta_1 = \atantwo(y_e \; , \; x_e) \\
    \beta_2 = \atantwo\left(-z_e \; , \; \frac{y_e}{\sin(\beta_1)} \right)
\end{cases}\enspace.
\end{equation}
Finally, we equate $\text{T}_e^2\left(\beta_3,\beta_4\right)=\text{T}_b^{2^{-1}}\text{T}_b^e$ to find $\beta_3$ and $\beta_4$ as
\begin{equation}\label{eq:beta34}
\begin{cases}
    \beta_3 = \atantwo\left(\text{T}_2^e(1,3) \; , \; -\text{T}_2^e(2,3) \right) \\
    \beta_4 = \atantwo\left(\text{T}_2^e(3,1) \; , \; \text{T}_2^e(3,2) \right)
\end{cases}\enspace.
\end{equation}
Due to the PM kinematics, the angles of the UJs are pairwise equal (i.e., $\beta_3=\beta_2$ and $\beta_4=\beta_1$). {These findings can be utilized for detecting non-idealities in the PM kinematics or for system calibration purposes.}

\begin{figure*}[!t]
    \subfloat[]{\includegraphics[width = 0.2\linewidth]{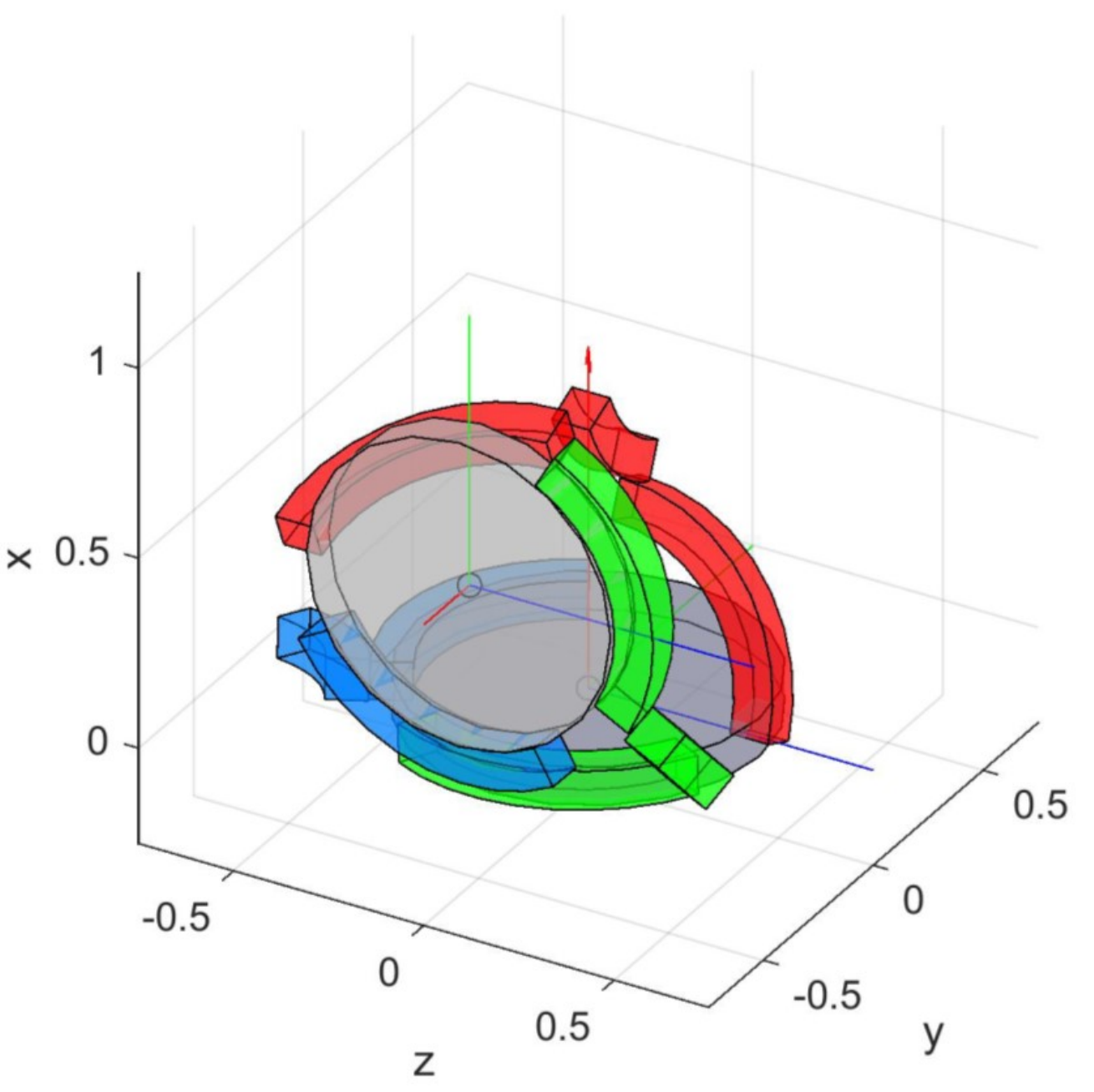}}
    \hfil    
    \subfloat[]{\includegraphics[width = 0.2\linewidth]{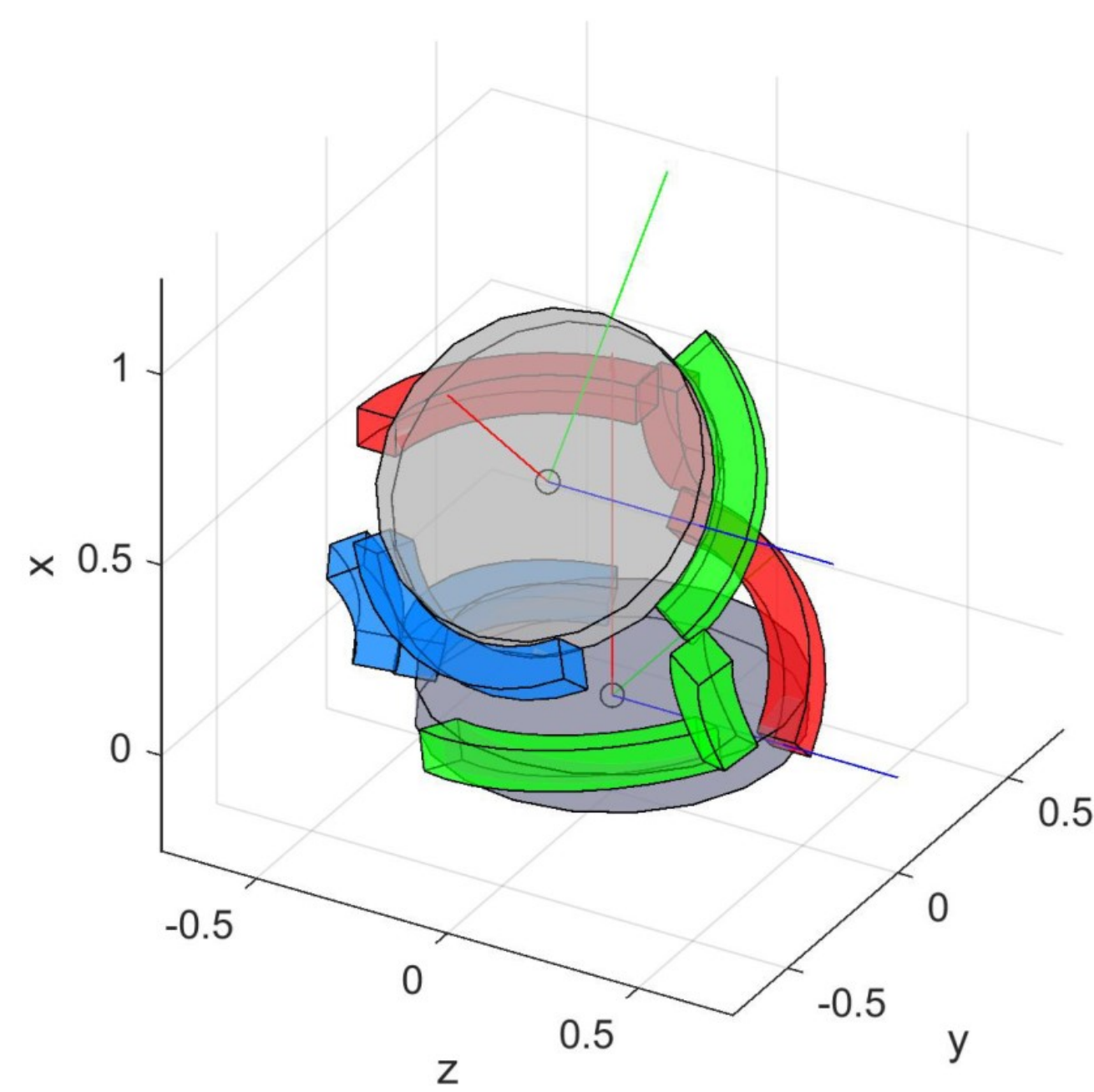}}
    \hfil
    \subfloat[]{\includegraphics[width = 0.2\linewidth]{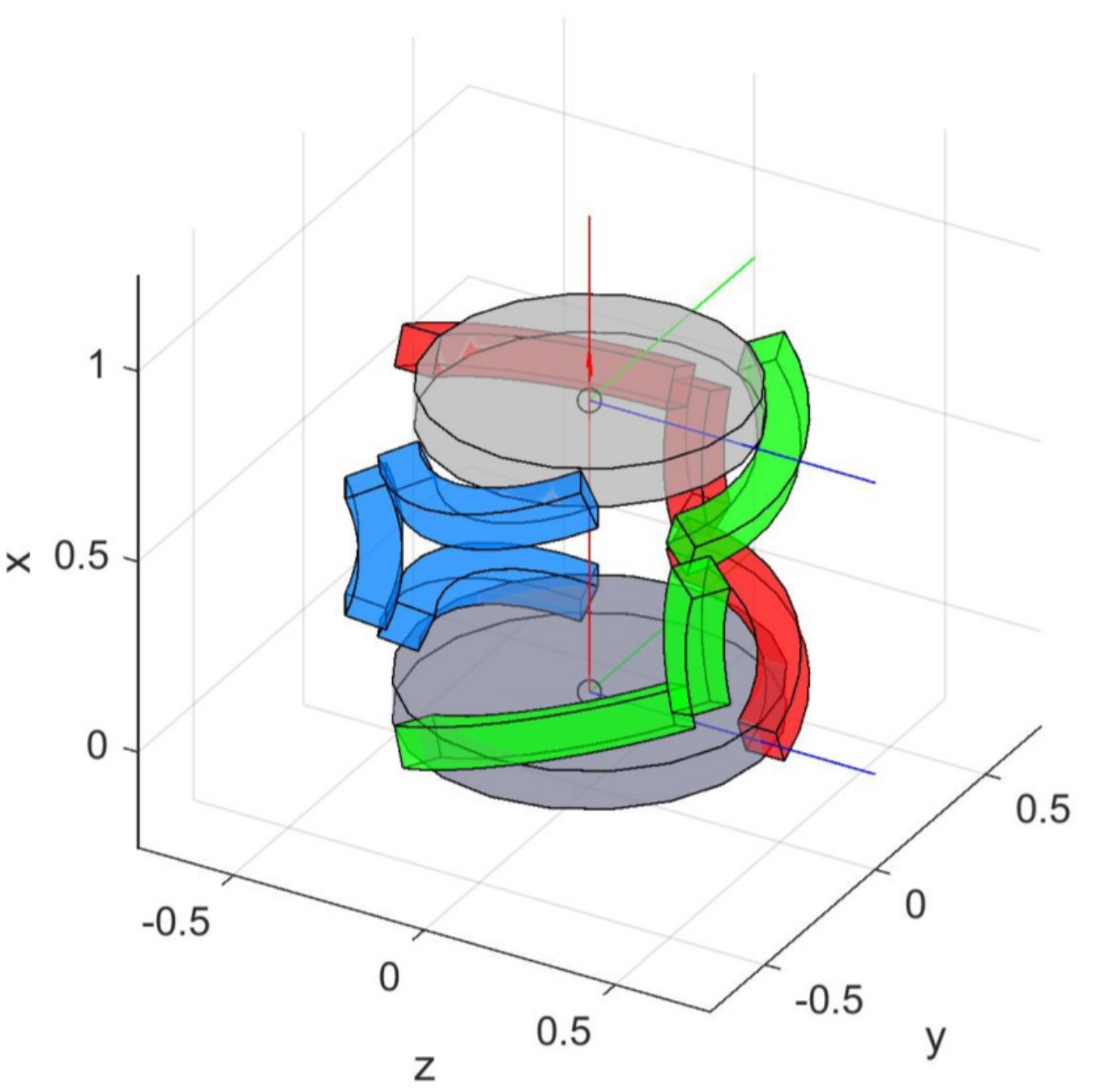}}
    \hfil
    \subfloat[]{\includegraphics[width = 0.2\linewidth]{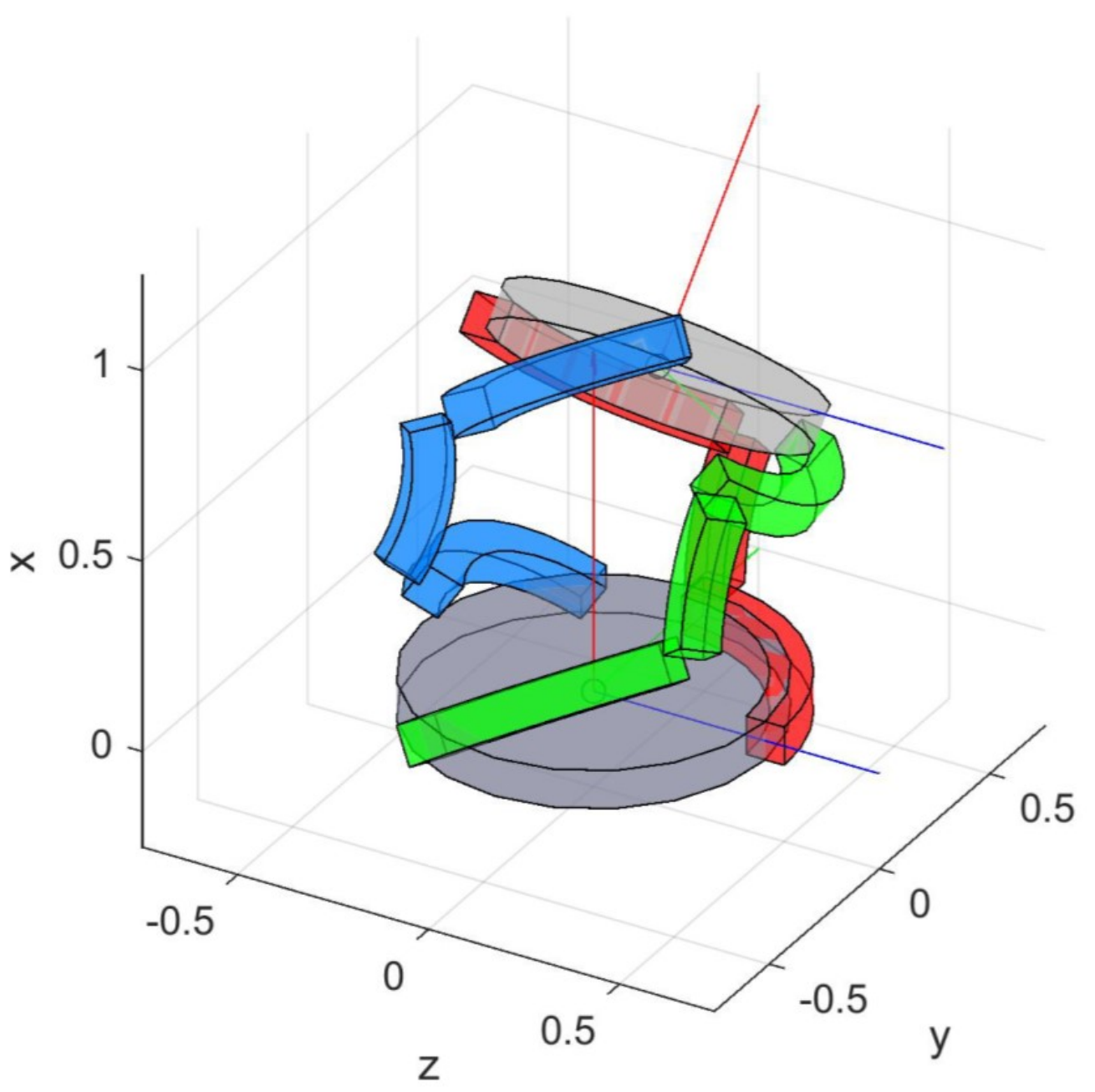}}
    \hfil
    \subfloat[]{\includegraphics[width = 0.2\linewidth]{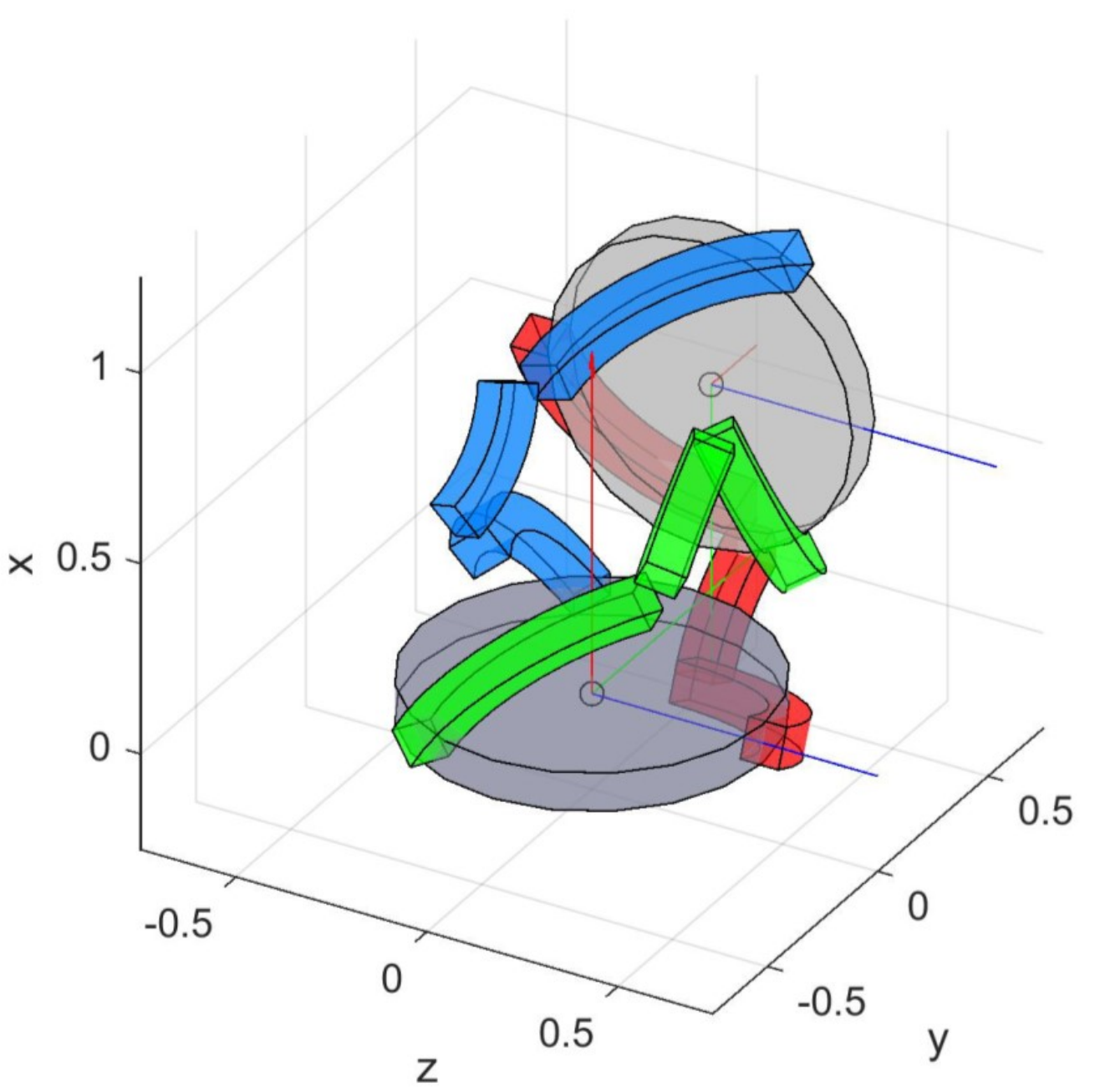}}
    \hfil
    \subfloat[]{\includegraphics[width = 0.2\linewidth]{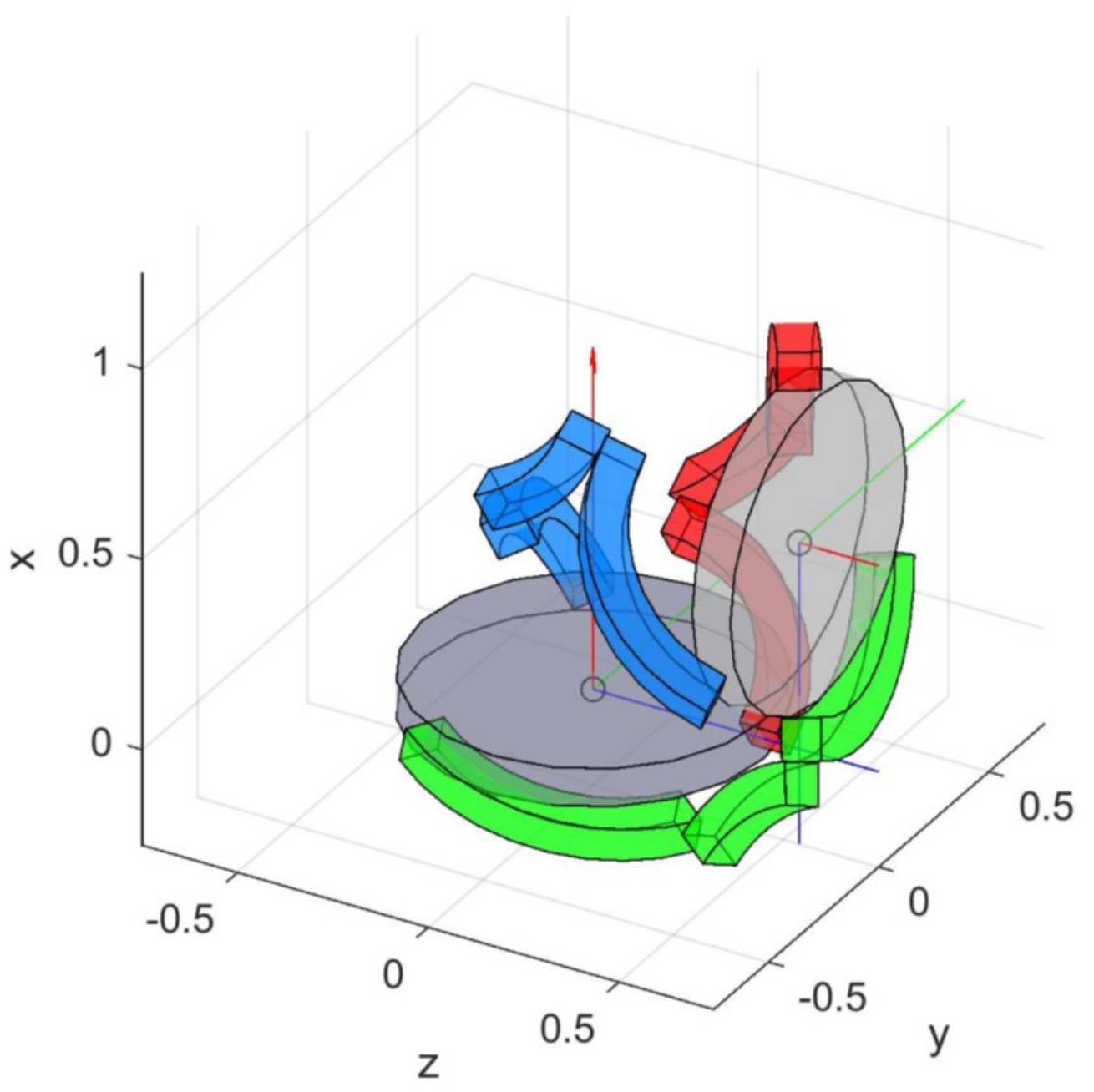}}
    \hfil
    \subfloat[]{\includegraphics[width = 0.2\linewidth]{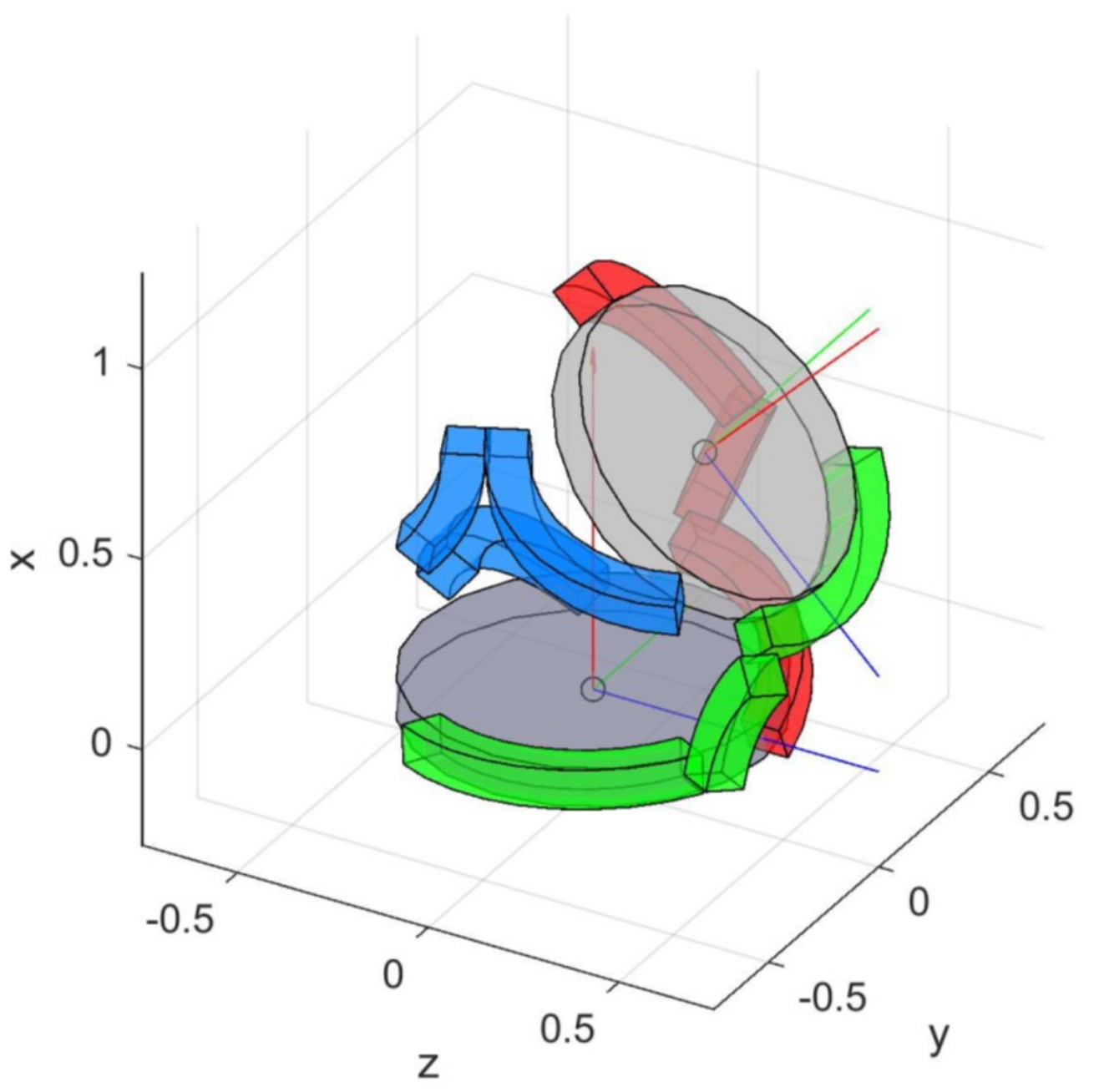}}
    \hfil
    \subfloat[]{\includegraphics[width = 0.2\linewidth]{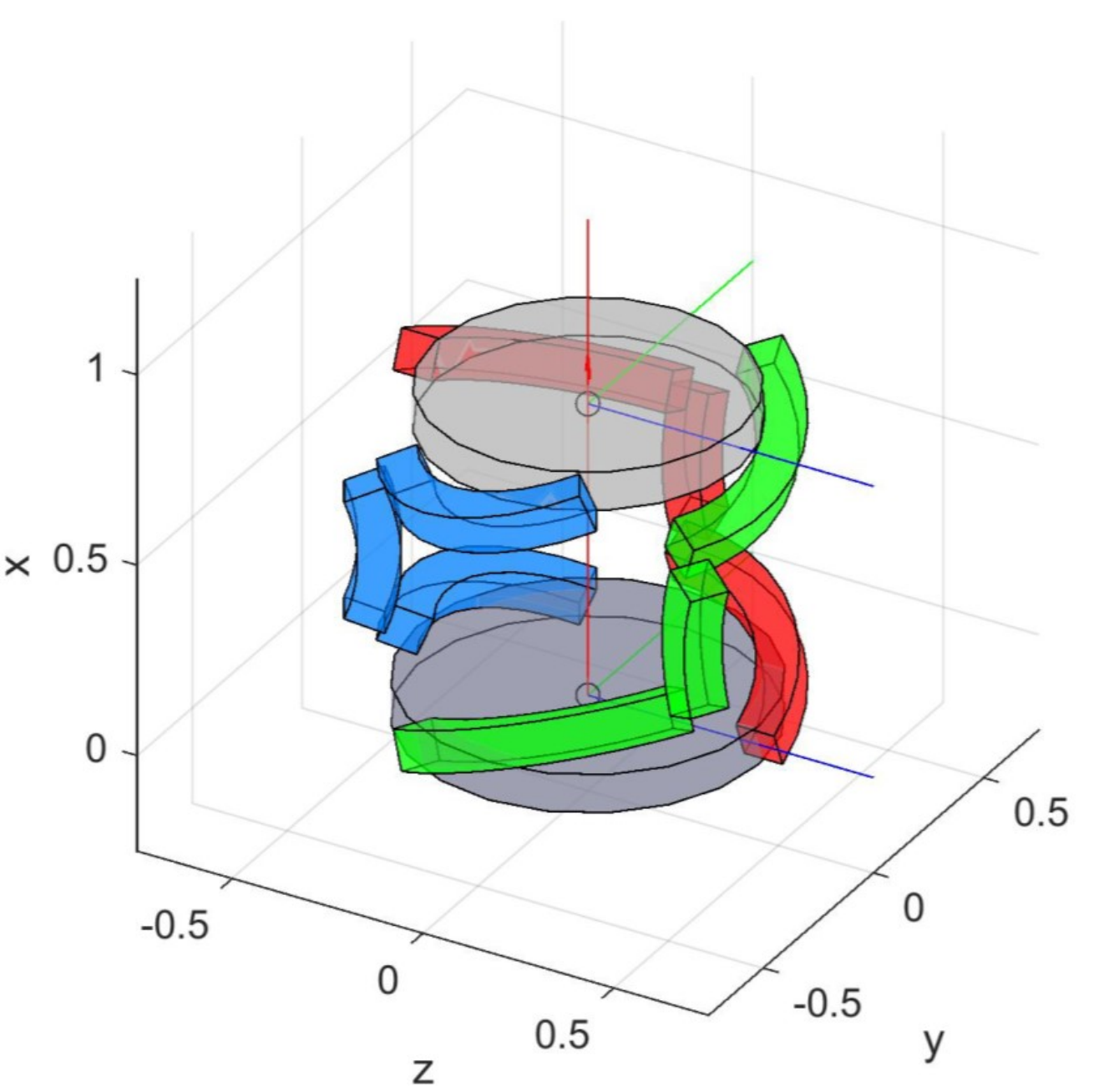}}
    \hfil
    \subfloat[]{\includegraphics[width = 0.2\linewidth]{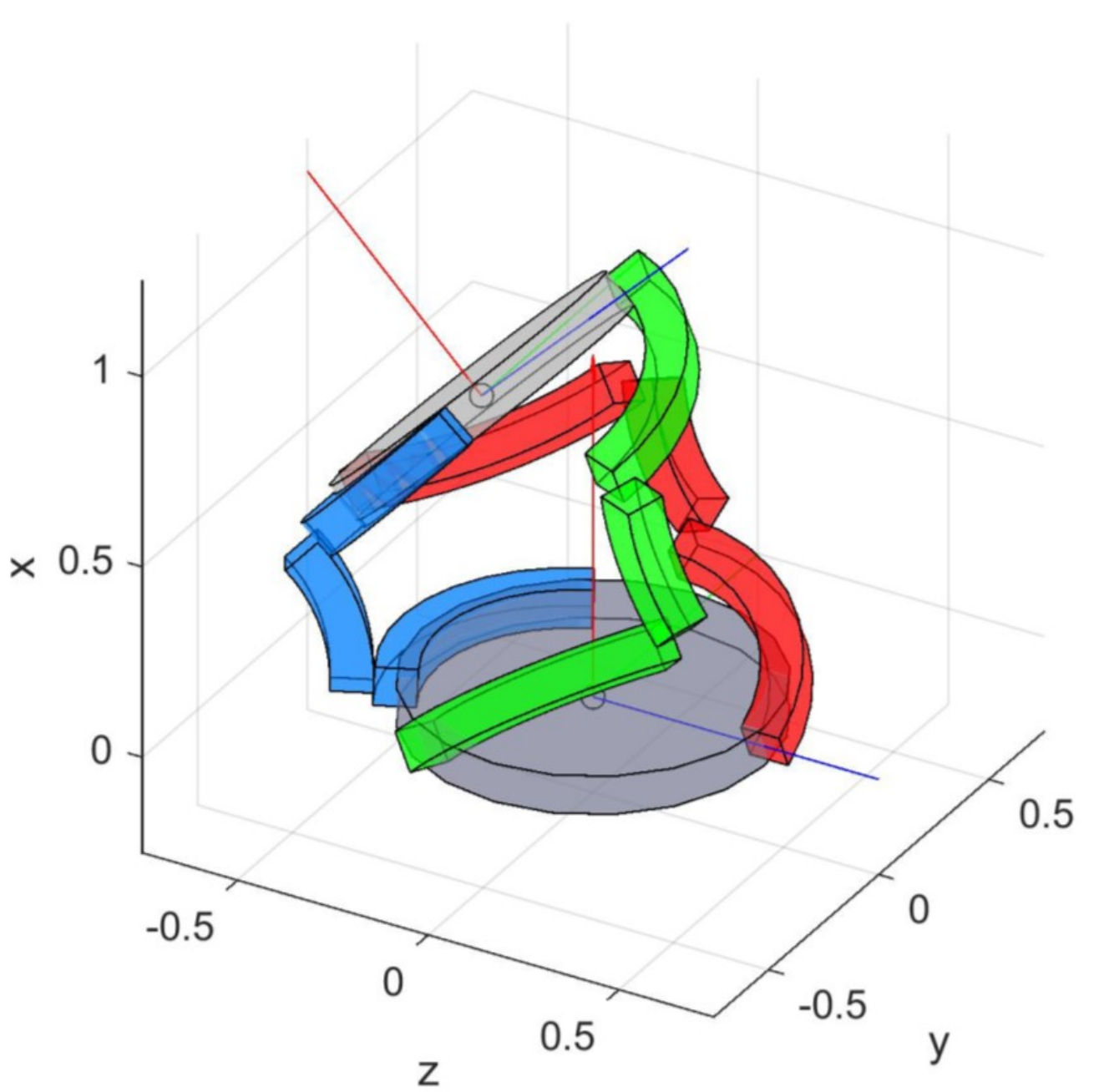}}
    \hfil
    \subfloat[]{\includegraphics[width = 0.2\linewidth]{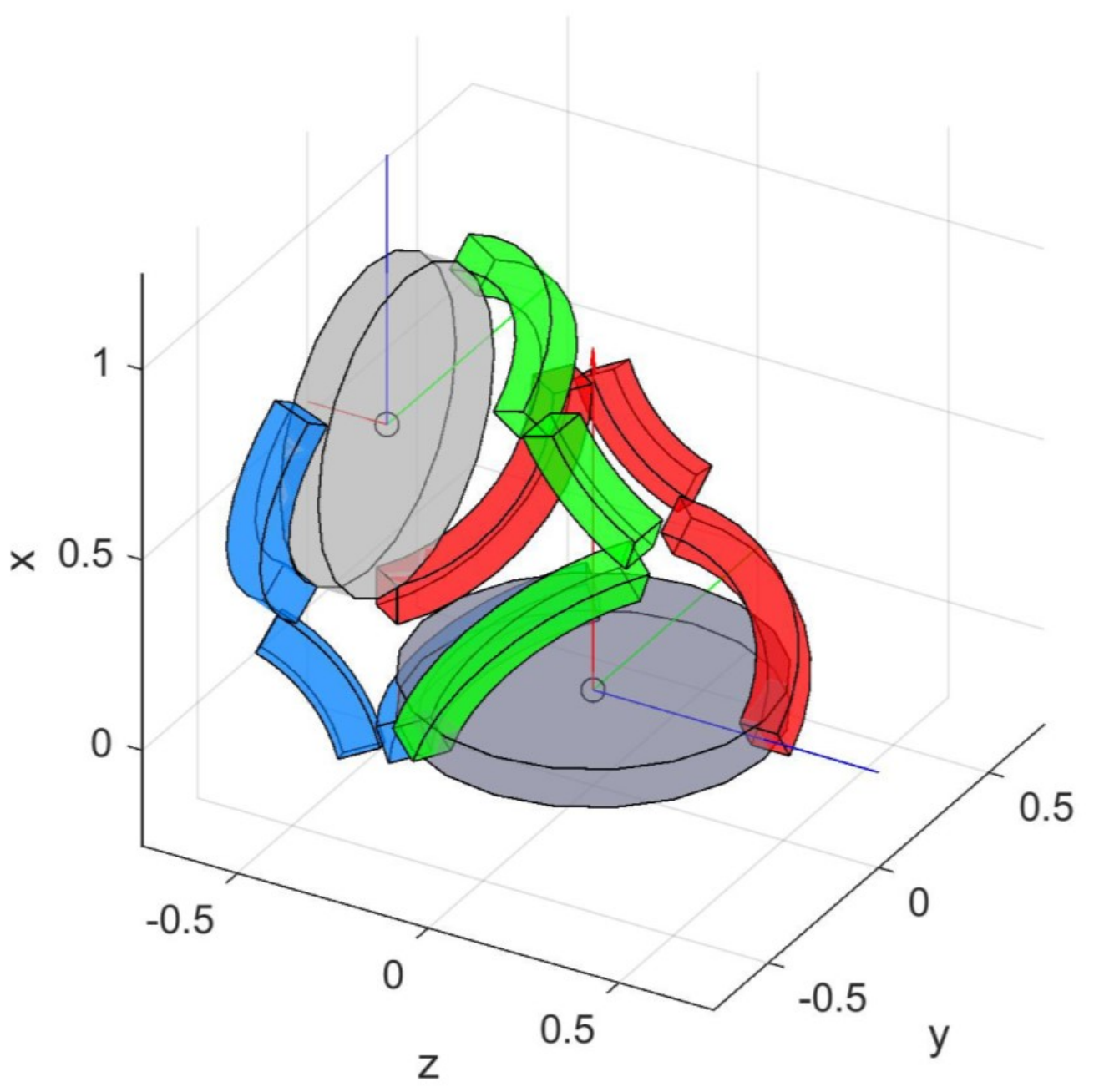}}
    \caption{{The various panels depict the nominal range of motion of the 2 DoFs VS joint, illustrating the kinematic behavior of the system using the inverse kinematic model. Panels (a) to (e) represent motion around the $Y_b$-axis, while panels (f) to (j) showcase a revolution around the $Z_b$-axis} (from $\frac{\pi}{2}$ to $-\frac{\pi}{2}$).}
\end{figure*}

\subsection{Single Leg Forward Kinematics}
The transformation matrix $\text{T}_b^e(q)$ that encodes the pose of the end-effector w.r.t. $\{S_b\}$ using joint angles $q$, must be equivalent to $\text{T}_b^e(\textbf{x})$, obtained through the Euler angles parametrization \textbf{x}. 

Equating these two matrices and solving for \textbf{x} yields
\begin{equation}
\begin{cases}
\label{eqn:FK_sol}
x_{e} = \text{T}_{1,4}(q)\\
y_{e} = \text{T}_{2,4}(q)\\
z_{e} = \text{T}_{3,4}(q)\\
\alpha_x = \atantwo\left(\text{T}_{3,2}(q),\text{T}_{3,3}(q)\right)\\
\alpha_y = \atantwo\left(-\text{T}_{3,1}(q),\sqrt{\text{T}_{3,2}(q)^2 + \text{T}_{3,3}(q)^2}\right)\\
\alpha_z = \atantwo\left(\text{T}_{2,1}(q), \text{T}_{1,1}(q)\right)  
\end{cases} \hspace{-1em},
\end{equation}
\noindent where $\text{T}_{i,j}(q)$ denotes the element of the $\text{T}_b^e(q)$ matrix at the ith row and jth column.
Therefore, given all the joint angles of a single leg, \eqref{eqn:FK_sol} provides the pose of the end-effector {w.r.t.} the fixed frame $\{S_b\}$.

\subsection{{Inverse Kinematics}}
Given the pose of the end-effector $\textbf{x}$, the IK of the PM yields the joint angles of a given leg. Exploiting the geometrical constraints 
\begin{equation}
\label{eqn:leg_constraint}
\begin{cases}
q_3 = q_2 + \pi\\
q_4 = -q_1
\end{cases} \enspace,
\end{equation}
{that} hold for the chosen mounting arrangement of the legs \cite{sofka}, and following the procedure detailed in \cite{lemerle2021wrist}, the IK solution is
\begin{equation}\label{eqn:IK_q1_sol}
\hspace{-0.9em}
\begin{cases}
    q_1 = \atantwo\left(\hspace{-0.5em}
    \begin{array}{l}
        x_e (1-c_{\alpha}) + (y_e c_{\eta}  + z_e s_{\eta})s_{\alpha} s_2 ,\\
        \quad d (1 - c_{\alpha}) (2 - (1 + c_{\alpha})c_2^2)
    \end{array}\hspace{-0.5em}\right) \vspace{0.5em} \\       
    q_2 = \acos\left(\frac{y_e \sin(\eta) - z_e \cos(\eta)}{d s_{\alpha}} \right) \hspace{6.8em},\\
    q_3 = q_2 + \pi \\
    q_4 = -q_1
\end{cases}
\end{equation}\hspace{-0.5em}
where $s_2$, $c_2$ {represent} $\sin(q_2)$, $\cos(q_2)$, and $s_{\alpha}$, $c_{\alpha}$ {represent} $\sin(\alpha)$, $\cos(\alpha)$. 
{The reader can find additional mathematical details about the computations in \cite{lemerle2021wrist}. The IK of the PM is subsequently utilized in Section~\ref{sec:pos_control} to determine the motor reference angles achieving a desired posture $u_r$.}

\subsection{Parallel Mechanism Forward Kinematics}
\label{subsect:Parallel_FK}
By equating the transformation matrices obtained using the joint angles of different legs, it is possible to derive all the joint variables of a single kinematic chain as a function of the sole sensorized joint angles, which are the first ones of each leg.
Computing $x_e$ and $y_e$ as functions of the joint angles of alternatively legs A and B yields
\begin{equation}
\label{eqn:FK_sensor}
\begin{cases}
    \text{T}_{1,4}(q_A) = x_e(q_A) = x_e(q_B) = \text{T}_{1,4}(q_B) \\
    \text{T}_{2,4}(q_A) = y_e(q_A) = y_e(q_B) = \text{T}_{2,4}(q_B)
\end{cases}
\enspace.
\end{equation}
Aligning the fixed base frame with the first local frame of leg A (hence $\eta_A = 0\text{, } \eta_B = \frac{2}{3}\pi\text{, and } \eta_C = \frac{4}{3}\pi$) and using \eqref{eqn:leg_constraint}, it is possible to solve the first equation of \eqref{eqn:FK_sensor} to obtain
\begin{equation} \label{eq:sinA2}
    s_{A_2} = \frac{(s_{B_1} - s_{A_1})(1 - c_{\alpha}) + s_{\alpha}c_{B_1}s_{B_2}}{s_{\alpha}c_{A_1}}  \enspace,
\end{equation}
where $s_{*_n}, c_{*_n}$ indicate the sine and cosine of the n-th joint angle of the leg $*$.
Substitute \eqref{eq:sinA2} into the second equation of \eqref{eqn:FK_sensor} to get
\begin{equation} \label{eqn:qb2_sol}
    \hspace{-0.5em}
    q_{B_2} = \atantwo\left(
    \begin{array}{l}
        \hspace{-0.5em}\sqrt{2}\left(s_{\alpha} t_5 - t_1 \sqrt{h_1 + h_2}\right) \enspace,\\
       \hspace{-0.5em} \quad t_1 (1 - t_3 - t_2 c_{\eta_B} + c_{A_1}s_{\eta_B}s_{\alpha})
    \end{array}\hspace{-0.5em}\right) \,, \\   
\end{equation}
where $t_1, t_2, t_3, t_4, t_5, h_1, h_2 \in \mathbb{R}$ are defined as 
\begin{equation}
    \begin{cases}
    t_1 = c_{\alpha} - 1\\
    t_2 = c_{A_1}c_{B_1}\\
    t_3 = 1 + c_{\eta_B}\\
    t_4 = c_{A_1}^2 + c_{B_1}^2 \\
    t_5 = c_{\eta_B}c_{A_1}s_{B_1} - c_{B_1}s_{A_1} \\
    h_1 = t_4 - t_2^2 ( 1 + c_{\eta_B}^2)\\
    h_2 = t_1 (1-t_2) - 2 t_2 t_3)c_{\eta_B}\\
    \end{cases} \enspace.
\end{equation}

Since the encoders measure $q_{A_1}$ and $q_{B_1}$, \eqref{eqn:qb2_sol} and \eqref{eqn:leg_constraint} yield all the joint angles of leg B, and \eqref{eqn:FK_sol} provides the pose of the end-effector \textbf{x} as a function of the sole sensorized variables.
{This strategy was employed to obtain real-time feedback on wrist posture, and its validation is presented in Section~\ref{sect:experimental_validation}.}

\subsection{Static Equilibrium}
\label{subsect:static_eq}
To solve the static equilibrium of the PM, we open the kinematic chains {at their last joints} and impose the equilibrium of the end-effector subject to the external wrench and the reaction forces from the coupling joints. {We then} enforce the equilibrium of each leg, considering the previously computed reaction forces.
This procedure, reported in detail in Appendix A, yields
\begin{subequations}\label{eqn:static_act}
    \begin{empheq}[left=\empheqlbrace]{align}
& \tau_a = J_a^{\top}(q)H^{\top}(q) \widetilde{w} \label{eqn:eq_a}\\
&      \tau_{\overline{a}} =  J_{\overline{a}}^{\top}(q) H^{\top}(q) \widetilde{w} = 0_{9\times1}  \quad,\label{eqn:eq_na}\\
&      w_e= -G(q)H^{\top}(q)\widetilde{w}  \label{eqn:eq_ee}  
    \end{empheq} 
\end{subequations}
\noindent where \eqref{eqn:eq_a} expresses the equilibrium of the actuated joints,  \eqref{eqn:eq_na} defines the balance of the torques on the non-actuated joints ({that} must be null), and  \eqref{eqn:eq_ee} imposes the equilibrium of the end-effector. 
As $GH^{\top}$ is a fat matrix with full row rank, all the possible solutions for \eqref{eqn:eq_ee} are expressed by
\begin{equation}\label{eq:w_tilde}
    \tilde{w} = -(GH^{\top})^R w_{e} + P \Lambda \enspace,
\end{equation}
\noindent where $(GH^{\top})^R$ is a right-inverse of $GH^{\top}$, the columns of P form a basis of the $\ker(GH^{\top})$, and $\Lambda$ is a vector of coefficients that parametrizes the constraint wrenches achieving the static equilibrium of the end-effector. 
Substituting \eqref{eq:w_tilde} into \eqref{eqn:eq_a} yields
\begin{equation}
    \label{eqn:sol_eq_a}
    \tau_a = -J_a^{\top} H^{\top} (GH^{\top})^R w_{e} + J_a^{\top} H^{\top} P \Lambda  \enspace,
\end{equation}
where $\Lambda$ must satisfy \eqref{eqn:eq_na}, leaving only one DoF $\lambda \in \mathbb{R}$ that modulates the internal forces of the 2 DoFs VS joint through the basis $N_0(u)$, defined in \cite{lemerle2021wrist}. 
Since the actuated torque acting on each leg is a non-linear function of the deflection $\delta = q_1 - \theta_m$ between the first joint and motor angles (i.e., $\tau_{a} = f(\delta)$), the stiffness of the elastic transmission $K(\delta) = -\frac{\partial \tau_{a}}{\partial \delta}$ varies with the deflection.
{Therefore, modulating the internal forces via the parameter $\lambda$ not only influences the active deflection of the elastic elements but also impacts joint stiffness. This strategy is employed in Section~\ref{sec:stiff_control} to control the stiffness of the device.}
{Solving \eqref{eqn:static_act} has been instrumental in sizing the actuation system and computing the maximum load the wrist can manipulate, as reported in Section~\ref{sect:discuss}, or determining the range of internal torques achievable with a lower load.}

\section{Elastic Transmission Mechanism}\label{sect:elastic_tr}
\begin{figure}
    \centering
        \includegraphics[width = 0.9\linewidth]{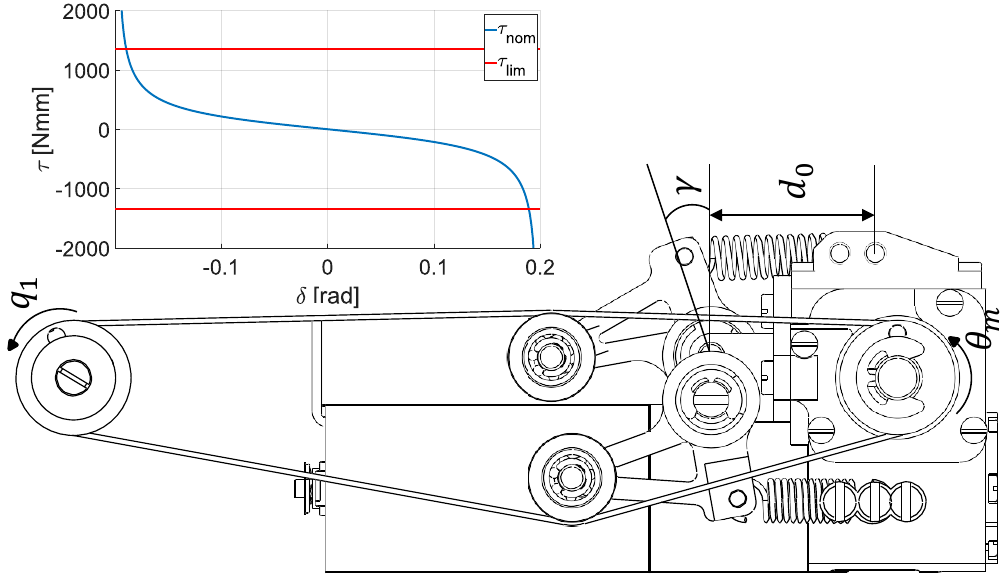} 
    \caption{{Schematic representation} of the non-linear elastic transmission mechanism. The {top-left plot} depicts the output torque {(blue line)} as a function of the deflection $\delta$ {within the nominal range of motor torque (red lines)}.}
    \label{fig:VSA_Scheme2}
\end{figure}

The elastic {transmission} mechanism, shown in Figure~\ref{fig:VSA_Scheme2}, employs an antagonistic setup of linear springs, tendons, and pulleys to achieve non-linearity thanks to their geometric configuration.
To derive the output torque delivered by the elastic transmission, first, we compute the elastic energy stored in the springs as a function of the lever angle $\gamma$, obtaining
\begin{equation}\label{eq:elastic_energy}
    U_{s}(\gamma) = \frac{1}{2}k(L(\gamma) - L_0)^2 \enspace,
\end{equation}
where
\begin{equation}
        L(\gamma) = \sqrt{(l_n \sin(\gamma) + d_0)^2 + (l_n(1-\cos(\gamma)))^2} \enspace,
\end{equation}
is the length of the spring as a function of $\gamma$, $L_0$ is the spring free length, $d_0$ is its length when $\gamma = 0$, and $l_n$ represents the length of the horizontal lever arm. 
Then, we differentiate \eqref{eq:elastic_energy} to obtain
\begin{equation}
    \label{eqn:tau_gamma}
   \tau(\gamma) = -\frac{\partial U_{s}}{\partial \delta} = -\frac{\partial U_{s}}{\partial \gamma} \frac{\partial \gamma}{\partial \delta} = -\frac{\partial U_{s}}{\partial \gamma} (\frac{\partial \delta}{\partial \gamma})^{-1} \enspace. 
\end{equation}
The relationship $\delta(\gamma, L_t)$ can be found by imposing the conservation of the tendon length $L_t$, as {outlined} in Appendix B. 
Therefore, given $\delta$ from the {encoder measurements} and assuming $L_t$ {is known by design, we can compute $\gamma$, and subsequently deduce the actuated torque.}

To {determine} the stiffness of the elastic transmission $K$, it is sufficient to differentiate \eqref{eqn:tau_gamma}, {yielding}
\begin{equation}
    K(\gamma) = -\frac{\partial\tau}{\partial\delta} = -\frac{\partial\tau}{\partial\gamma} (\frac{\partial \delta}{\partial \gamma})^{-1} \enspace.
\end{equation}

\section{Hardware Description}\label{sect:hardware}
\subsection{Mechanical Hardware}
\begin{figure}[!t]
        \subfloat[]{\label{fig:vsw_fullarch}\includegraphics[width = 0.95 \linewidth]{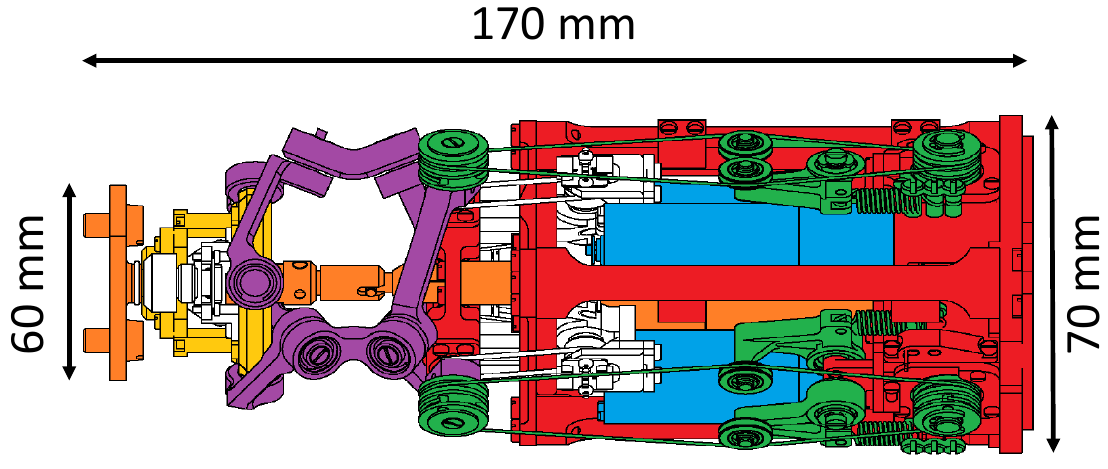}}
        \vfil
        \subfloat[]{\label{fig:vsw_leg_subsystem}\includegraphics[width = 0.52 \linewidth]{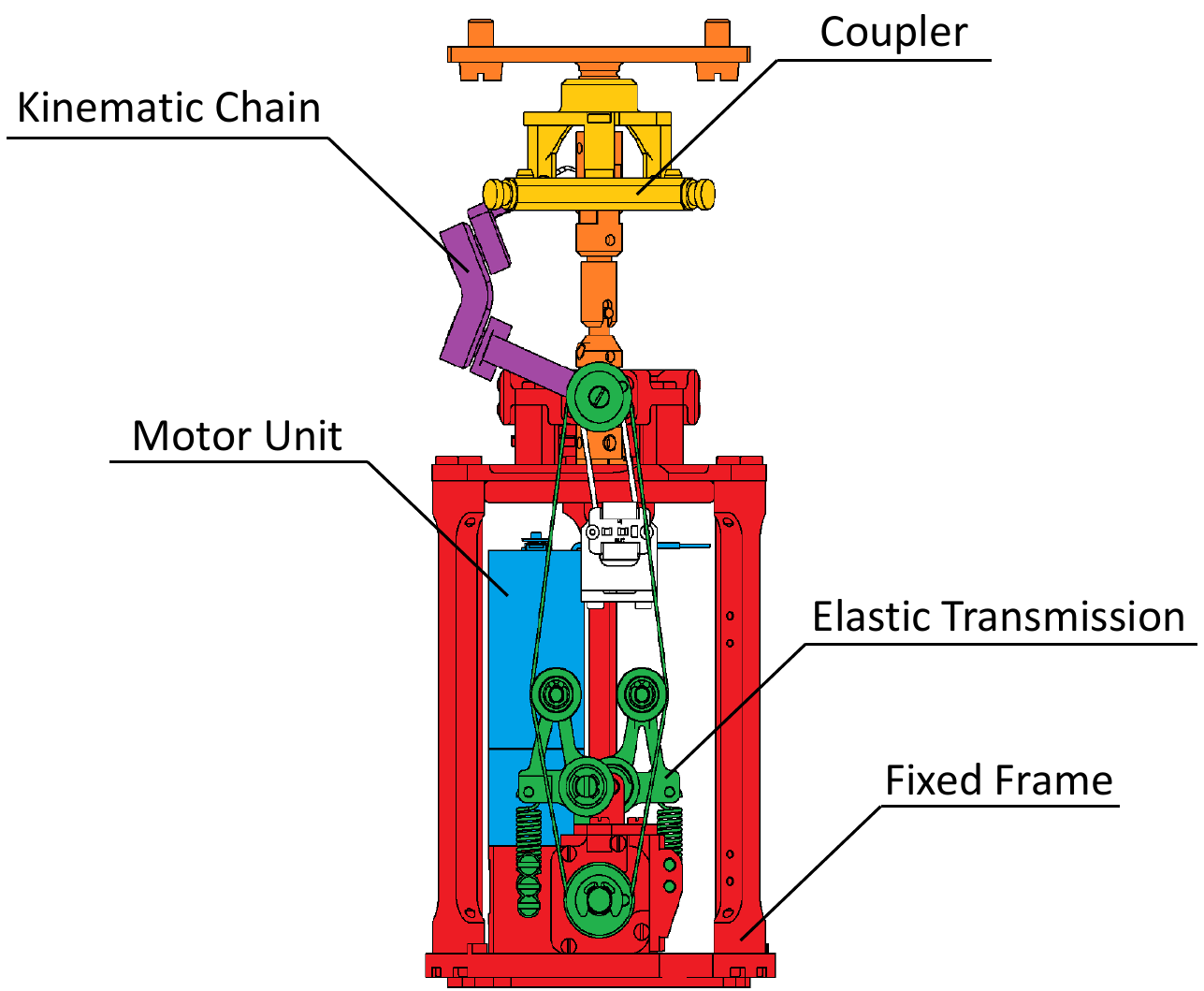}}
        \hfil
        \subfloat[]{\label{fig:vsw_ps}\includegraphics[width =0.45\linewidth]{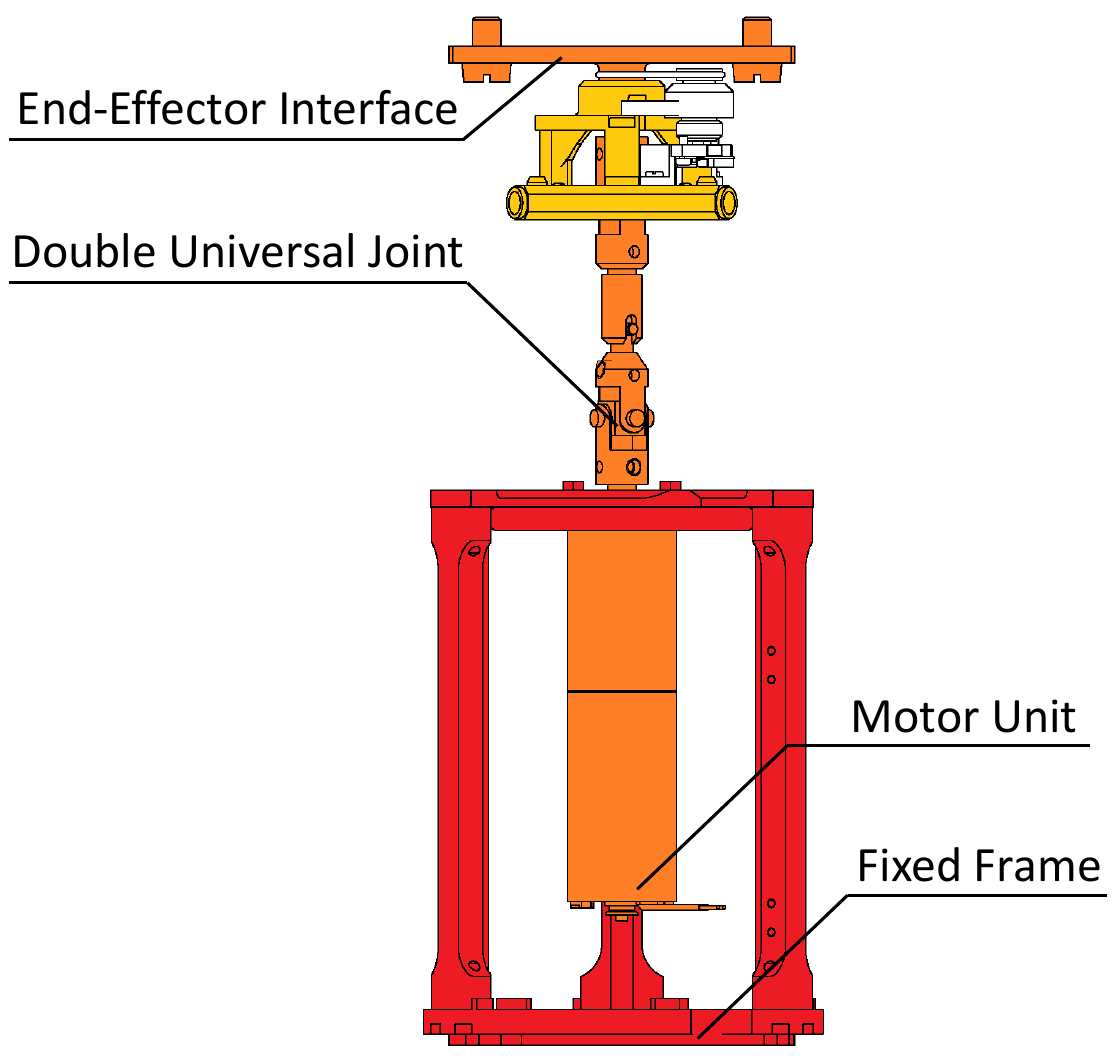}}
    \caption{CAD of the VS-Wrist. Each color highlights a different macro-component: red for the fixed frame, blue for the 2 DoFs VS joint motor units, green for the elastic transmission mechanism, magenta for the kinematic chains, orange for the PS unit, and yellow for the coupler. Panel (a) offers a comprehensive 3D view of the device along with its dimensions. Panel (b) details the implementation of one leg and its VS unit, while panel (c) showcases the PS unit.}
    \label{fig:Vsw_Arch}
\end{figure}

Figure~\ref{fig:Vsw_Arch} shows the CAD of the VS-Wrist {along with} its overall dimensions. The fixed frame, depicted in red, has a diameter of 70 mm, while the {coupler}, shown in yellow, has a diameter of 60 mm. {The total length of the device is 170 mm, and its weight is 1110 g. These dimensions align with the average sizes of a human wrist and forearm as per \cite{nasa_dim} and \cite{damerla2022design}.}

The VS joint features three identical sides composed of the serial arrangement of a motor unit (in blue), a non-linear elastic transmission (in green), and a kinematic chain of four non-coplanar revolute joints (in magenta).
The motor unit comprises a Maxon DCX19S~GB~KL 12V motor, its associated gearbox, and a low-efficiency worm gear transmission to ensure the non-backdrivability of the system.
Thanks to this design choice, the springs can apply constant forces to maintain high stiffness levels or balance static loads without power consumption.
The resulting transmission ratio is 420, with an efficiency of 0.28. The maximum continuous torque deliverable to the gearbox output shaft is 1351 Nmm, while the maximum intermittent torque is 3883 Nmm. The nominal power consumption of each motor is 16 W, but since the actuation is non-backdrivable, the power {demand} is negligible when the wrist is not moving actively.

\begin{figure}[!t]
    \centering
    \subfloat[]{\label{fig:FEM_displacement}\includegraphics[width = 0.48\linewidth]{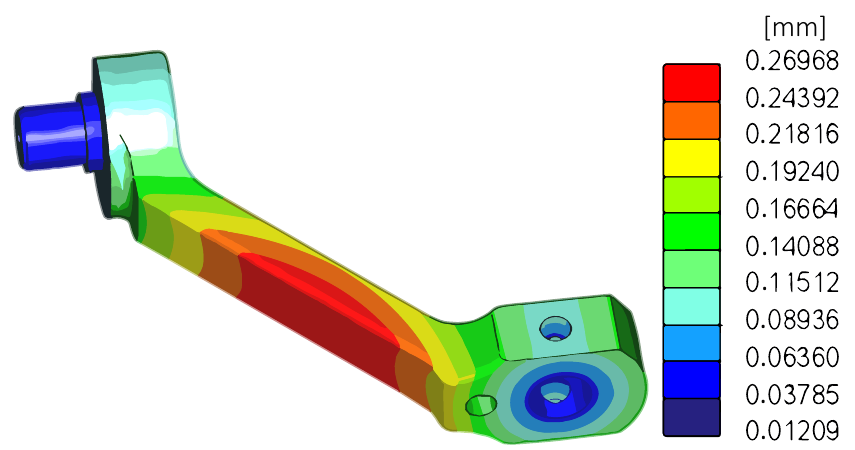} }
    \hfil
    \subfloat[]{\label{fig:FEM_vonMises}\includegraphics[width = 0.48\linewidth]{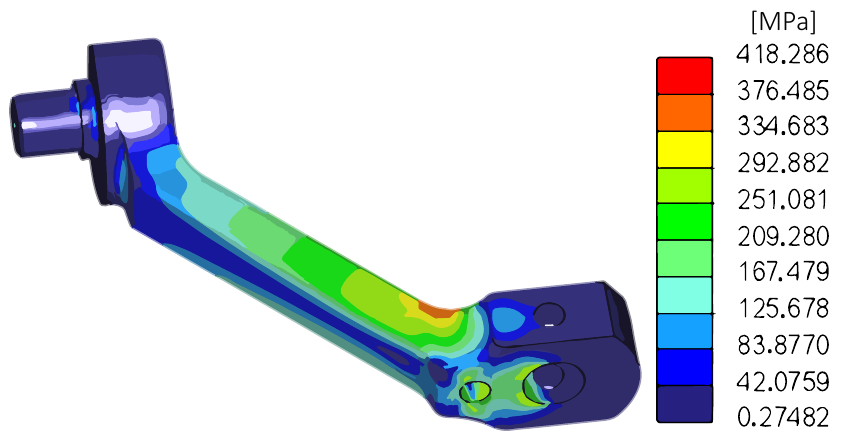}}
    \caption{FEM analysis of the first link of the parallel manipulator under maximum load condition. Panel (a) reports the magnitude of the displacement, and panel (b) shows that the von Mises stress remains below the yield strength of the material ($\sigma_Y = 503$ MPa).}
    \label{fig:FEM}
\end{figure}

The kinematic chains{, made of aluminum alloy Ergal 7075-T6, evenly} distribute around the fixed frame. To assess the robustness of the device, we performed a FEM analysis of the kinematic chains under maximum load conditions. Figure~\ref{fig:FEM} {illustrates} the magnitude of the displacement and von Mises stress of the first link when static equilibrium is reached under an external weight of 5.2 kg. Figure~\ref{fig:FEM_displacement} shows that the maximum displacement of the link is restrained and approximately located in the middle section. Figure~\ref{fig:FEM_vonMises} demonstrates stress concentration in a rounded corner and {affirms} that the kinematic chain withstands the maximum load condition {as} the highest stress remains below the yield strength of the material $\sigma_Y = 503$ MPa.
The characteristic length $d = 49$  mm and angular deviation $\alpha = \frac{\pi}{4}$ are {carefully selected} to prevent internal collisions between different legs {and attain the desired workspace. The impact of these design parameters on the system workspace and compliance was previously examined in \cite{lemerle2021wrist}. Additionally, \cite{chang2022use} provides a manipulability analysis assessing wrist dexterity while using a terminal device.}

The elastic transmission mechanism achieves VS by adjusting the spring preload. For this purpose, one end of each spring is fixed to the frame, while the other end attaches to the lever arm mechanism that regulates the tension of the tendons.
The elastic elements {employed} are the extension springs T31320 from MeterSprings, and the tendons are crafted using Liros DC120 from Unlimited Rope Solutions. 
A traction machine pulled the cables at 90\% of their maximum tolerated tension to ensure that their length remains consistent during the regular operation of the wrist.

The PS actuator features a planetary gearbox with a transmission ratio of 168:1, efficiency of 0.65, maximum continuous of 1204 Nmm, maximum intermittent torque of 2000 Nmm, and {a} power consumption of 16 W. 
Figure~\ref{fig:vsw_ps} illustrates the mechanical implementation of the PS unit. The double UJ structure compensates for parallel misalignment and non-constant velocities between the driving and driven shafts. {Importantly, the transmission is independent of the VS unit, meaning this motor does not modify the stiffness of the joint.}

\subsection{Electronic Hardware}
The device {incorporates} the NMMI electronics {detailed} in \cite{della2017quest}.
\indent Four independent PID controllers regulate the position of the DC motors, relying on feedback from four rotary magnetic encoders that measure the angle $\theta$ of the gearbox output shafts. Three additional encoders measure the first joint angle $q_1$ of each kinematic chain. The 12-bit programmable encoders AS5045 from Austriamicrosystems {provide} angular positions with a resolution of {0.09°}. Four motor drivers MC33887 from NXP Semiconductors independently drive the DC motors.
The current flowing to the motor is regulated at 1 kHz by modulating the duty cycle of a 20 kHz PWM signal.
The Programmable System-on-Chip (PSoC) 5LP CY8C58LP from Cypress manages the PID controller at a frequency of 1 kHz. The embedded microcontroller is a 32-bit Arm Cortex-M3 core plus DMA.
A daisy-chain connection allows various boards to communicate and share power from a single DJI TB47 (4500mAh) battery. 
The PSoC connects to the computer via micro-USB, and {communication is handled through the RS485 serial protocol.}

\section{Control Architecture}\label{sect:control}
\begin{figure}[!t]
\centering
\includegraphics[width=\linewidth]{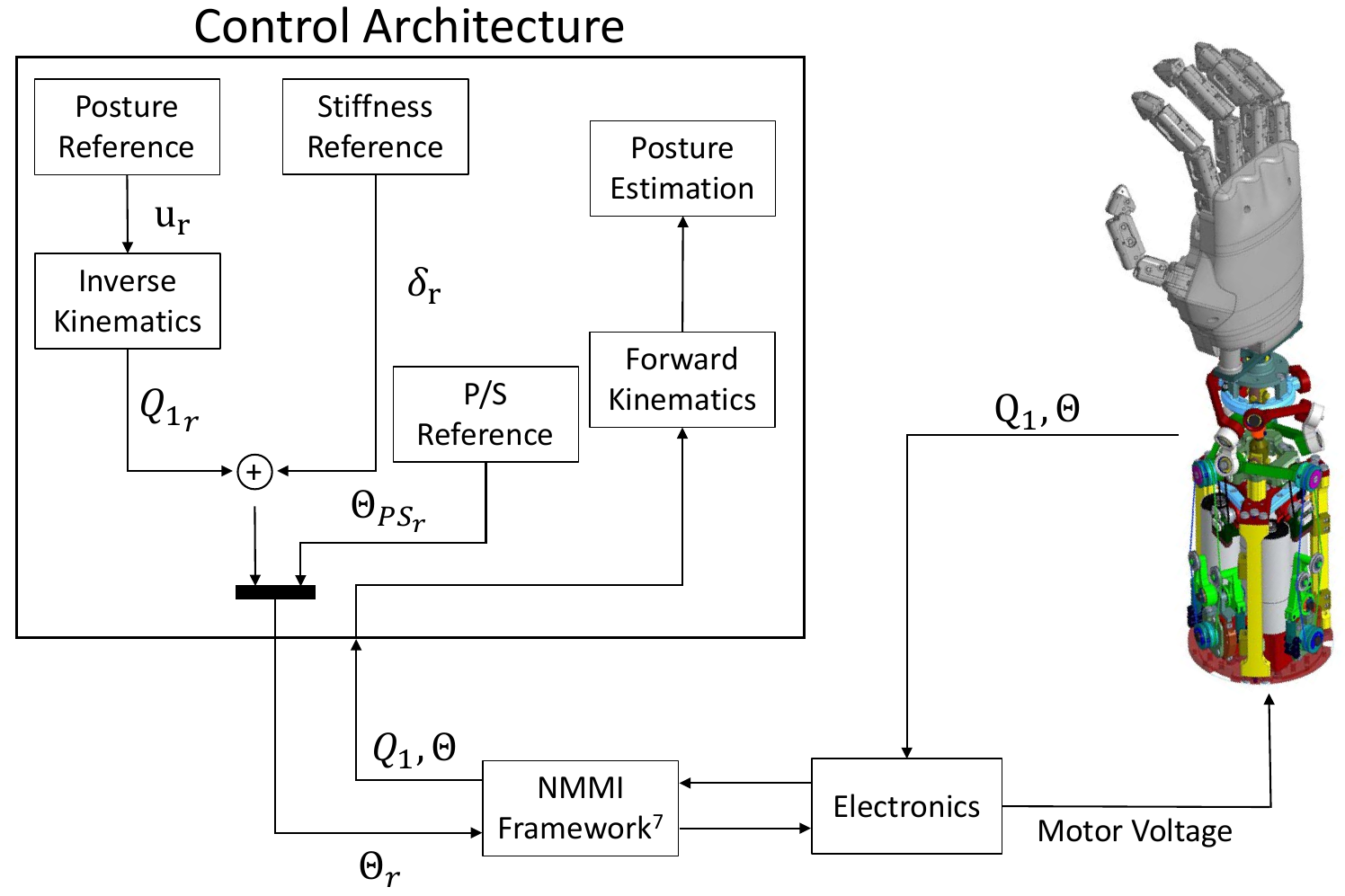}
\caption{Scheme of the control architecture. A Simulink model sets the position and stiffness references in real-time, communicating with the electronic hardware through the NMMI framework. Bidirectional communication ensures that encoder measurements are available to the user within the Simulink environment.}
\label{fig:Control_Architecture}
\end{figure}
To validate the effectiveness of the proposed hardware without {compromising} the inherent compliance of the system \cite{della2017controlling}, we opt for a simple open-loop controller. This control strategy regulates the motor positions {based} solely on kinematic considerations{, remaining} independent of the characteristics of the non-linear elastic transmission. The control architecture, schematized in Figure~\ref{fig:Control_Architecture}, {consists of} a position controller and a stiffness controller, {both} decoupled and {operating} in parallel. 

\subsection{Posture Control}\label{sec:pos_control}
The posture controller {capitalizes on the insight that the system naturally tends towards static equilibrium by minimizing the elastic energy stored in the springs}. In the absence of external or internal loads, this equilibrium occurs when the deflection $\delta$ is null. Consequently, the angular position of the first joint follows the corresponding motor angle. 

Given a desired pose of the wrist in minimum parametrization $u_{r}$, \eqref{eqn:IK_q1_sol} provides the corresponding first joint angle of each leg $Q_{1r} \in \mathbb{R}^3${. Consequently}, the reference position of the motors $\Theta_{r}^p \in \mathbb{R}^3${, is determined by}
\begin{equation} \label{eq:thetaref}
    \Theta_{r}^p = Q_{1r} = \mathcal{IK}_1(u_{r}) \enspace.
\end{equation}

Since the serial motor unit directly acts on the PS of the wrist, this DoF is independently regulated using a conventional proportional control loop.

\subsection{Stiffness Control}\label{sec:stiff_control}
As described in Section~\ref{subsect:static_eq}, the internal forces delivered by the actuation system $\tau_{s} = \lambda N_0(u)$ {govern} stiffness regulation. 
Although the base of the actuated internal torques {relies} exclusively on the kinematics of the manipulator, deriving the motor angles that {yield} these torques requires perfect knowledge of the elastic transmission model.
However, non-modeled phenomena, such as hysteresis, tendon stretching, and manufacturing variability, limit the reliability of the derived model.

For simplicity, we assume all the VS units to be identical, and we neglect the variations of $N_0$ with posture, considering it constant at the value achieved when the wrist is in the central position. 
Then, we refine this approximation through a calibration procedure described in Section~\ref{subsect:stiffcal}. We define the motor references that modulate the stiffness of the coupler $\Theta_r^s \in \mathbb{R}^3$ as
\begin{equation}
    \Theta_{r}^s = \delta_r N_0 = \delta_r \begin{bmatrix}
        1\\
        1\\
        1
    \end{bmatrix} \enspace ,
\end{equation}
where $\delta_r \in \mathbb{R}$ represents the desired deflection of each VS unit. 
Since the output stiffness monotonically increases with $\delta_r$, we adopt this variable {as a means to} control the stiffness.

\subsection{Combined Control}
Integrating stiffness and posture control, the device moves to the commanded posture while the VS units maintain the desired preload.
Due to the non-backdrivability of the actuation, external wrenches acting on the wrist do not influence the motor output shaft. 
As a result, the device {consistently displaces according to the commanded stiffness, leveraging} the embedded elasticity.
To concurrently manage both joint stiffness and position, we assume the two controllers to be perfectly decoupled. Consequently, we define the motor reference angles as:
\begin{equation}
        \Theta_{r}^c  = \mathcal{IK}_1(u_{r}) + 
        \delta_{r} \begin{bmatrix}
        1\\
        1\\
        1
    \end{bmatrix}\enspace.
\end{equation}
Finally, a simple proportional control yields the commanded motor angle.

\subsection{Hardware Implementation}\label{sect:VS-Wrist_HI} 
The hardware implementation of the high-level controller, schematized in Figure~\ref{fig:Control_Architecture}, was {conducted} in MATLAB Simulink using the NMMI framework \cite{della2017quest}. This setup enables real-time bidirectional communication between Simulink and the electronic components of the VS-Wrist. 
{To account for delays arising from the communication protocol and computations, the frequency of the Simulink diagram was reduced to 200 Hz, ensuring real-time execution.}

\section{System Calibration and Characterization}\label{sect:calibration}
We calibrate the system to address errors arising from manufacturing tolerances and other un-modeled phenomena.
The limitations of the prototype, {stemming from non-nominal behavior, are summarized } and discussed in Section~\ref{sect:prot_limit}.

\subsection{Experimental Setup}\label{subsect:expSetup}
We measure the posture of the wrist using a Vicon\endnote{Vicon, Motion Capture System. \url{https://www.vicon.com/}. Last Accessed: May 2023} motion capture system consisting of 12 optical cameras and 14 reflective markers.

\begin{figure}[!t]
    \centering
    \subfloat[]{\label{fig:MarkerSetOverlay} \includegraphics[height = 13 em]{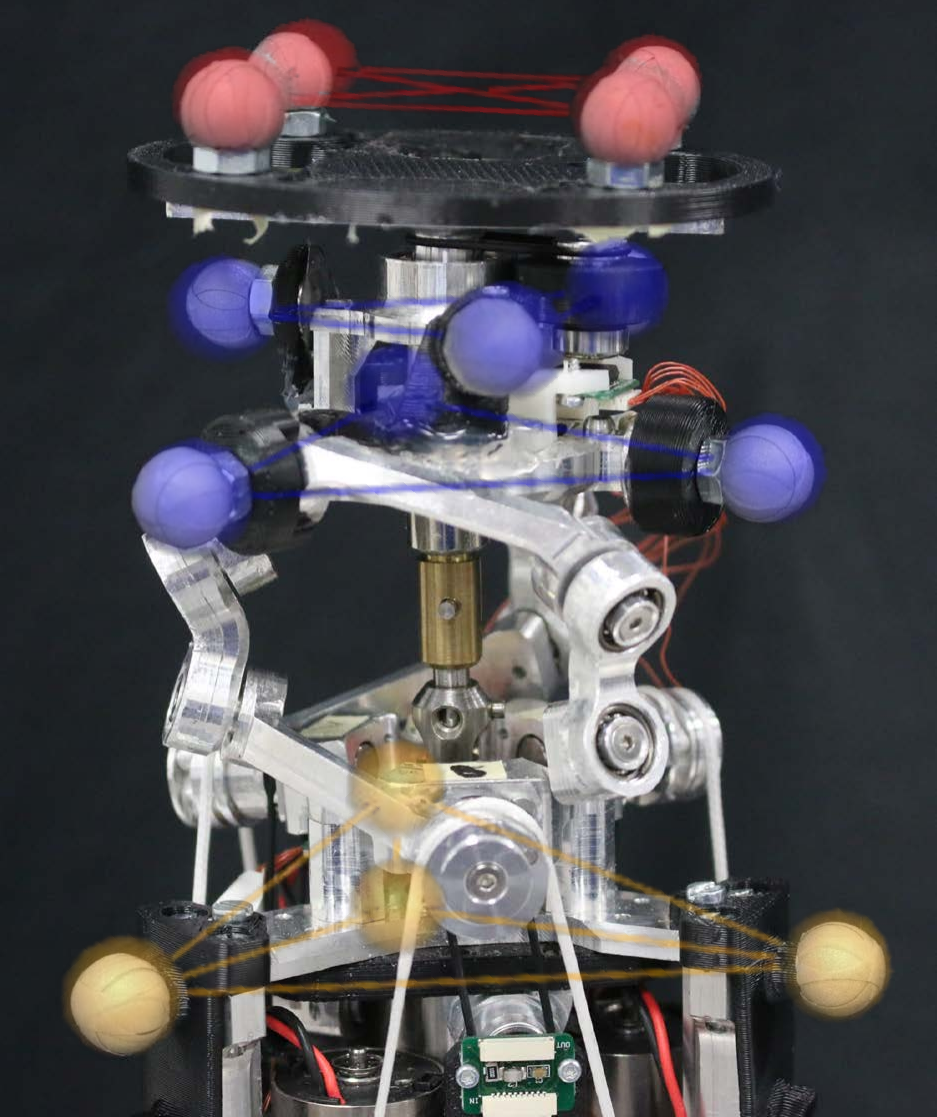}}
    \hfil
    \subfloat[]{\label{fig:MarkerObject} \includegraphics[height = 13 em]{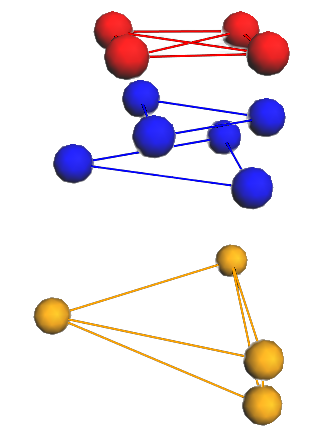}}
        \caption{Panel (a) shows the employed marker set, and panel (b) displays its reconstruction conducted by the motion capture system within the Vicon Nexus environment.}
    \label{fig:Markers}
\end{figure}

{With the aid of custom 3D-printed supports, three sets of markers are affixed to} the base frame, the coupler, and the end-effector, respectively (see Figure~\ref{fig:Markers}).
The camera data are sampled at a frequency of 100 Hz and exported to Matlab for post-processing.
{Utilizing} the marker trajectories, we reconstruct the absolute position and orientation of the reference frames $\{S_b\}$ and $\{S_e\}$ to compute the posture of the end-effector \textbf{x}, and the transformation matrix $\text{T}_b^e(\textbf{x})$ as in \eqref{eq:pose_euler_extend}. The experimental procedure {draws inspiration from} \cite{grioli2015variable} and \cite{mihcin2021wearable}.

\subsection{Stiffness Control Calibration}\label{subsect:stiffcal}
\begin{figure}
    \centering
    \includegraphics[width = 0.95 \linewidth]{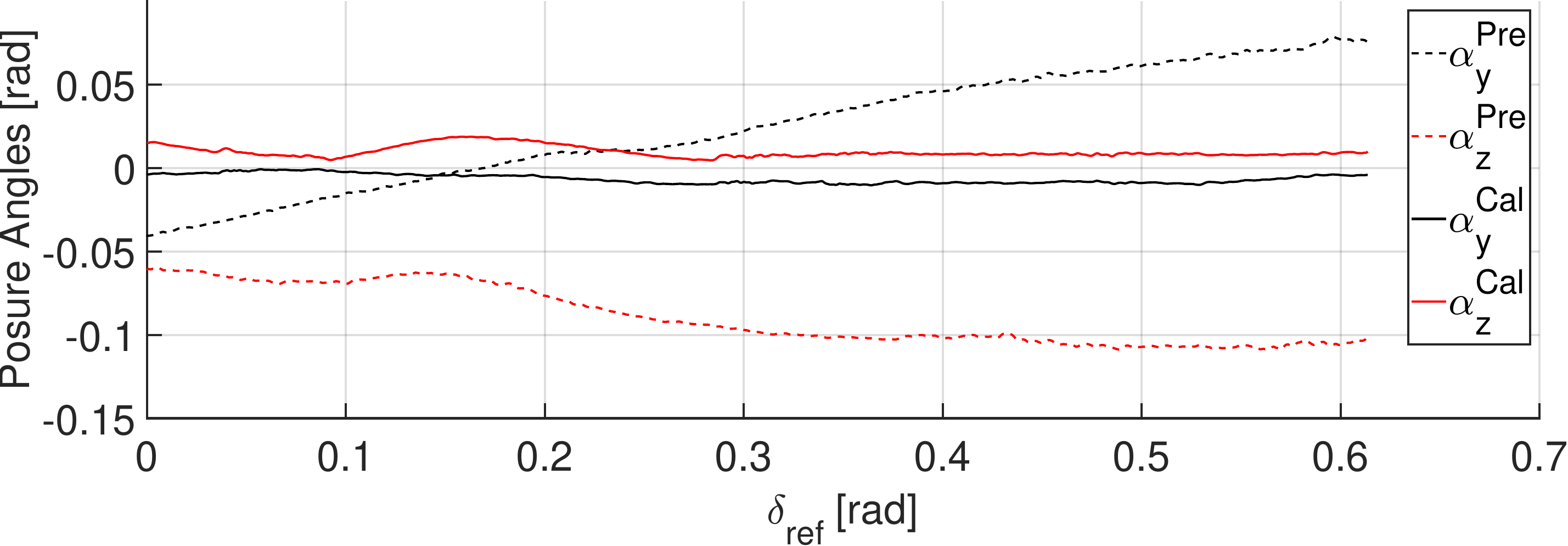}
    \caption{Calibration of the stiffness controller. The graph demonstrates that, prior to calibration, a pure stiffness command induces changes in the wrist posture due to manufacturing variability (dashed lines). Repeating the experiments after applying the compensation command validates that the proposed calibration process effectively mitigated this undesired phenomenon (continuous lines).}
    \label{fig:StiffnessCalibration}
\end{figure}

Although the system can theoretically modulate its stiffness without moving the end-effector, we observe a slight deviation of the coupler when controlling the prototype based on the nominal model (see Figure~\ref{fig:StiffnessCalibration}, dashed lines).
{To address this issue}, we fit the experimental data to {determine} a compensation command that mitigates the observed posture drift {using} least squares optimization. 
{Subsequently, we adjust} the reference posture angles as
 \begin{equation}
     \begin{cases}
        \alpha_{yc} = \alpha_{yr} + {P}_{y}^4(\delta_{r})\\
        \alpha_{zc} = \alpha_{zr} + {P}_{z}^4(\delta_{r}) 
    \end{cases} \, ,
 \end{equation}
where $\alpha_{yc},\alpha_{zc}$ represent the compensated angles derived from the original references $\alpha_{yr},\alpha_{zr}$ and ${P}_{y}^4(\delta_{r}), {P}_{z}^4(\delta_{r})$ are fourth-order polynomials of the variable $\delta_{r}$ that provide the compensation command.

\subsection{Combined Control Calibration and Characterization}\label{subsect:com_controlCal}
To characterize the behavior of the posture controller, we begin by testing each DoF separately using step and sine wave posture references. Subsequently, we combine all DoFs to estimate the RoM of the device. Each trial is repeated three times for each of the six tested stiffness configurations.

\begin{figure}
    \centering
    \subfloat[]{\label{fig:sine_unloaded}\includegraphics[width = 0.48 \linewidth, trim={0 34 0 30 em},clip]{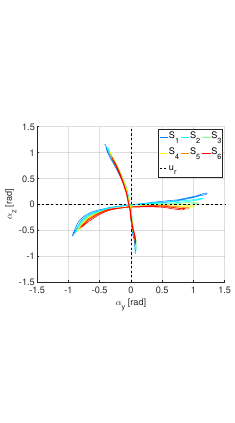}}
    \hfil
    \subfloat[]{\label{fig:GridUnloaded}\includegraphics[width = 0.48 \linewidth]{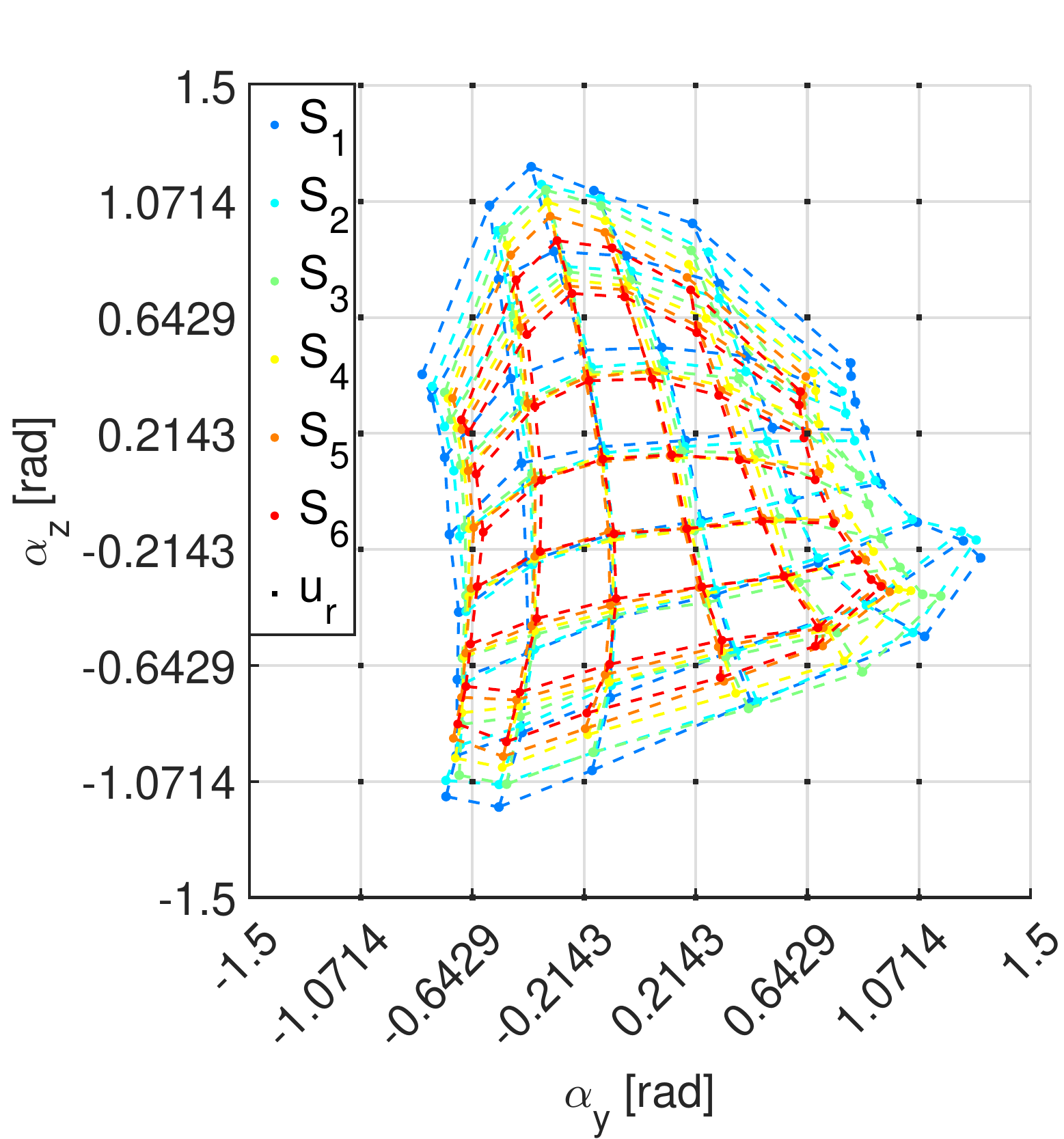}}
    \caption{Characterization of the kinematic behavior of the device. Each color represents a different stiffness configuration (increasing from $S_1$ to $S_6$). Panel (a) shows the device tracking sinusoidal references, acting on a single axis per time. Panel (b) shows the experimental RoM of the prototype obtained by sampling its 2D workspace with a grid.}
    \label{fig:posture_control}
\end{figure}

Figure~\ref{fig:sine_unloaded} {demonstrates how the device tracks} a sinusoidal reference at different stiffness configurations. {Notably,} the RoM of the device slightly reduces as the stiffness grows. At maximum stiffness, the device achieves $\pm 0.85$ rad for $\alpha_y$, and $[-0.8, 0.9]$ rad for $\alpha_z$. The proportional controller provides swift and accurate pursuit of motor reference, with an average $RMSE(\theta) = 0.02$ rad. However, the effective posture of the prototype {deviates from the sinusoidal reference by approximately} 40\% ($RMSE(u) = 0.35$ rad), {exhibiting undesired cross-axis movements.}

We sampled the 2D-RoM of the VS joint with a grid and acquired static measurements of the device posture, combining movements on both planes at various stiffness configurations. Figure~\ref{fig:GridUnloaded} portrays the RoM of the prototype along the FE and RUD DoFs, showing that the workspace of the device is smaller than the nominal one and reduces as the stiffness of the wrist {increases}. 
The average standard deviation of the posture among different repetitions of the same trial is very low ($\sigma_{u} = 0.04$ rad), {indicating} precise and reproducible control. 
{However, accuracy is compromised ($RMSE(\alpha_y) = 0.25$ rad, $RMSE(\alpha_z) = 0.50$ rad) due to the use of open-loop control based on the nominal model, resulting in significant discrepancies between the actual and controlled positions.}

Based on the identified kinematic behavior, we {developed} a linear map that adjusts the posture reference by leveraging experimental data. Given a desired posture $\overline{u}~=~[\overline{\alpha}_y \enspace \overline{\alpha}_z]^\top$ {within} the RoM of the device, we sample its four neighbors $\mathbb{U}^N_4(\overline{u})$ on the feedback posture grid. Subsequently, we compute a weighted average of their corresponding command to obtain
\begin{equation}\label{eq:uref_comp}
       \overline{\alpha}_\ast^c = \sum_{i = 1}^{4}\left(\frac{w_{\ast i}}{\sum_{i = 1}^{4} w_{\ast i}} \frac{\overline{\alpha}_\ast}{\alpha_{\ast i}}  \overline{\alpha}_{\ast i} \right)   \quad \text{for $\ast = y, \, z$},
\end{equation}
where $\alpha_{\ast i} \in \mathbb{U}^N_4(\overline{u})$ represents a neighbor sampled on the experimental grid, $\overline{\alpha}_{\ast i}$ denotes its associated commanded posture, $w_{\ast i} = (\alpha_{\ast i} - \overline{\alpha}_\ast)^{-2}$ weights its distance from the desired posture, and $\overline{u}^c = [\overline{\alpha}_y^c \enspace \overline{\alpha}_z^c]^\top$ is the compensated posture reference. We account for the variability of the kinematic behavior with the stiffness by computing $\overline{u}^c$ on the two grids whose stiffness is the closest to the reference configuration and then performing a weighted average of the results based on the difference in stiffness from the desired one.

\subsection{Posture Reconstruction Calibration}\label{subsect:posrecon}
We evaluated the accuracy of the posture reconstructed by the encoders along the grid, considering the measurements of the motion capture system as ground truth.
The previous experiment {demonstrated} that the posture along one DoF is influenced by the posture along the {other} and the stiffness. Therefore, we adapted the method in \cite{iosa2014assessment} by using multivariate linear regression to fit the measurements of the motion capture system as
\begin{equation}
    \begin{cases}
        \alpha_y^{GT}=p_{y0}+p_{yy}\alpha_y^e+p_{yz}\alpha_z^e+p_{ys}\delta_r + \varepsilon_y\\
        \alpha_z^{GT}=p_{z0}+p_{zy}\alpha_y^e+p_{zz}\alpha_z^e+p_{zs}\delta_r + \varepsilon_z
    \end{cases}
\end{equation}
\begin{figure}[!t]
    \centering
    \subfloat[]{\label{fig:alpha_y_ReconstructGridFit}\includegraphics[width = 0.72 \linewidth]{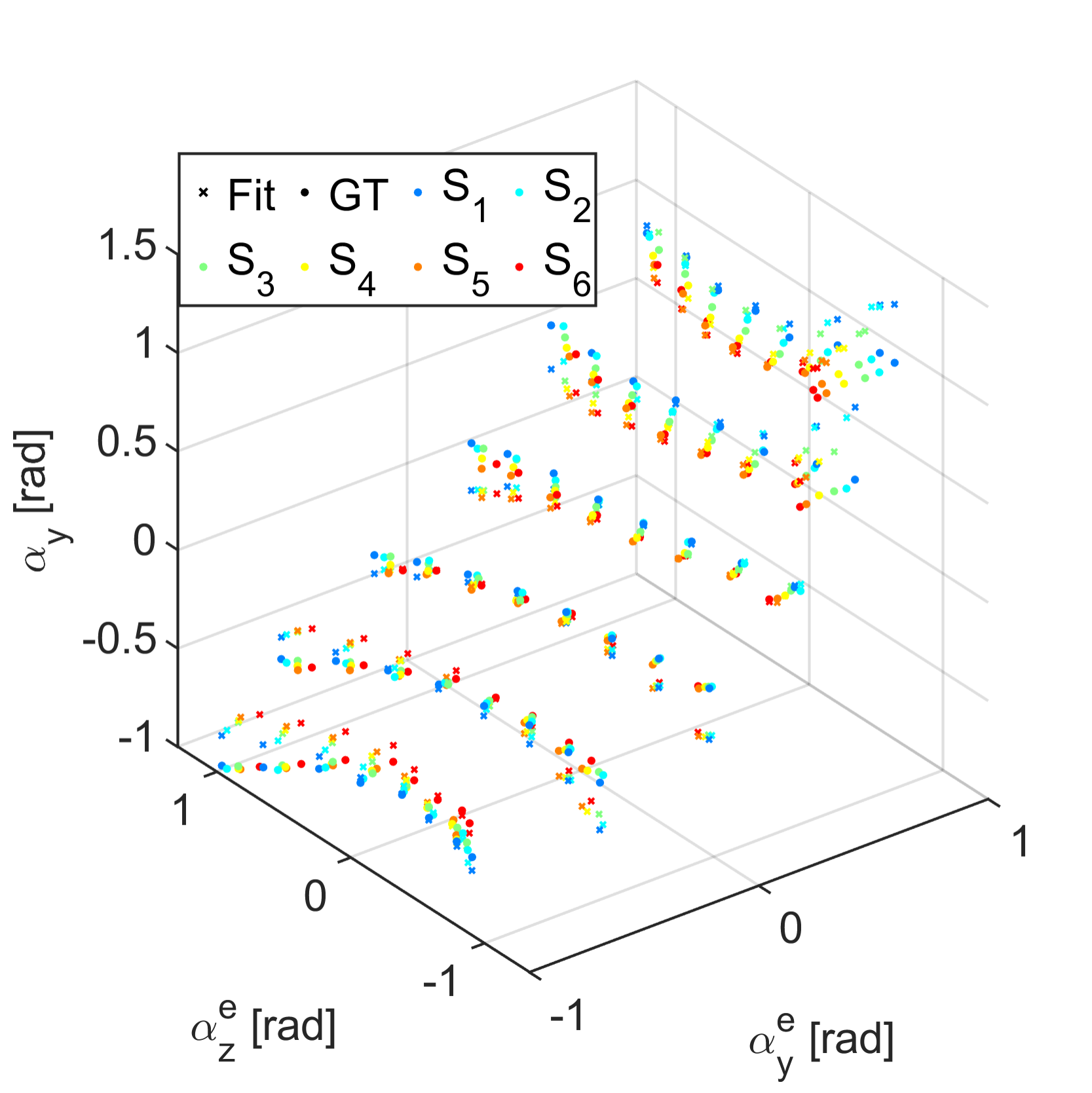}}\\
    \subfloat[]{\label{fig:alpha_z_ReconstructGridFit}\includegraphics[width = 0.72 \linewidth]{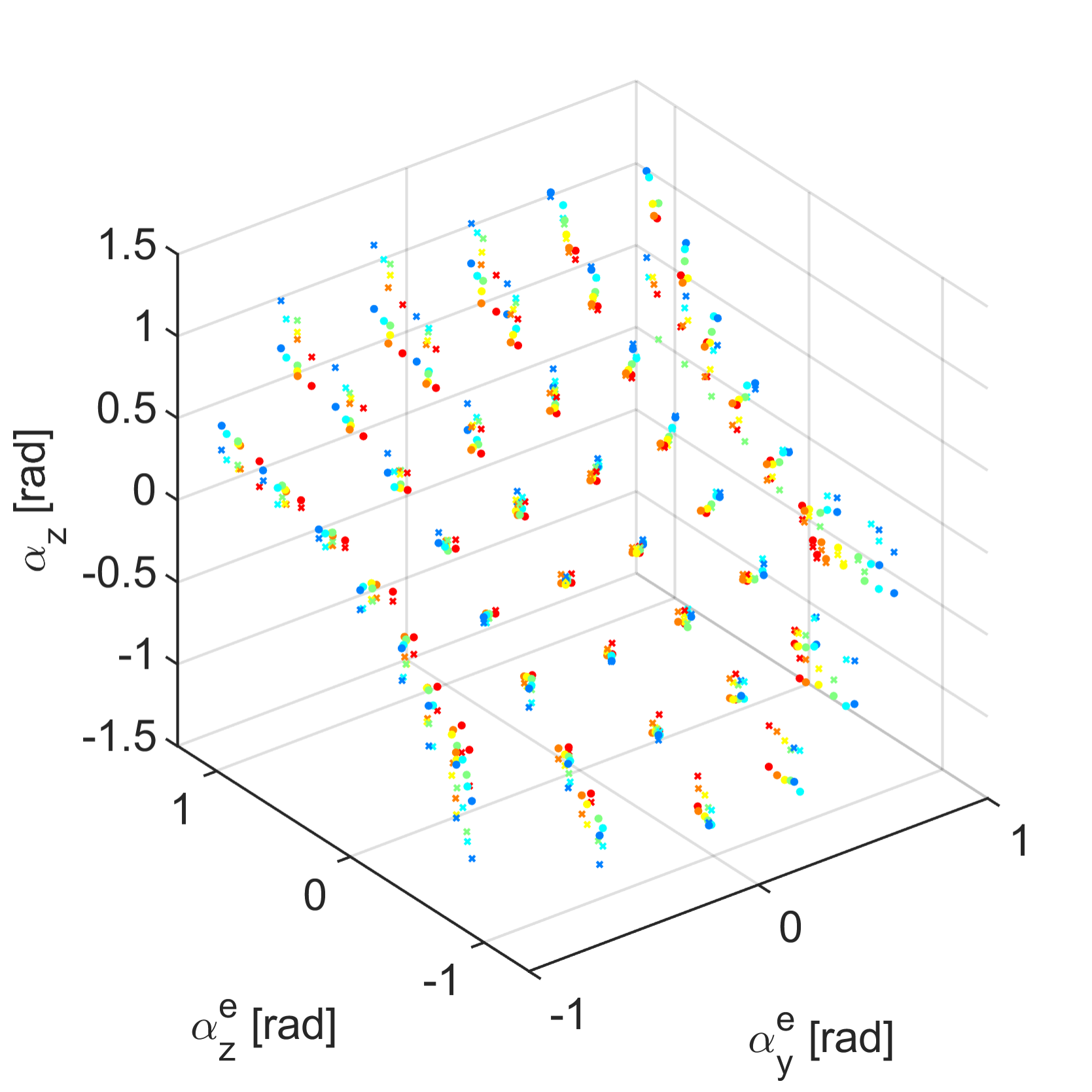}}
    \caption{Wrist posture reconstructed through multivariate linear regression. Each color corresponds to a distinct stiffness configuration (increasing from $S_1$ to $S_6$). Crosses indicate the posture reconstructed by the algorithm, while dots represent the posture measured with motion capture, assumed as the ground truth.} 
    \label{fig:PostureReconstruction}
\end{figure}
\begin{table}[!t]
\small\sf\centering
\caption{Coefficients and statistics from the multivariate linear regression used to estimate wrist posture.  Inside the parentheses is reported the 95\% confidence interval of the estimated parameters.}
\label{tab:posture_reconstruction}
\begin{tabular*}{\linewidth}{@{\extracolsep{\fill}}c | c c c c c c}
\toprule
 $\alpha_\ast$ & $p_{\ast 0}$ & $p_{\ast y}$ & $p_{\ast z}$ & $p_{\ast s}$ & $RMSE$ & $R^2$\\
\midrule
$\alpha_y$ & \begin{tabular}{@{}c@{}}0.06 \\ \tiny{(0.03, 0.09)}\end{tabular}  &
\begin{tabular}{@{}c@{}}0.91 \\ \tiny{(0.89, 0.94)}\end{tabular} &
 \begin{tabular}{@{}c@{}}-0.15 \\ \tiny{(-0.18, -0.14)}\end{tabular} & 
 \begin{tabular}{@{}c@{}}-0.09 \\ \tiny{(-0.17, -0.01)}\end{tabular} & 0.12 & \small{0.96}\\
 $\alpha_z$ & \begin{tabular}{@{}c@{}}\small{-0.03} \\ \tiny{(-0.08, 0.01)}\end{tabular}  &
\begin{tabular}{@{}c@{}}\small{0.15} \\ \tiny{(0.11, 0.18)}\end{tabular} &
 \begin{tabular}{@{}c@{}}\small{0.75} \\ \tiny{(0.72, 0.78)}\end{tabular} & 
 \begin{tabular}{@{}c@{}}\small{-0.08} \\ \tiny{(-0.21, 0.04)}\end{tabular} & \small{0.18} & \small{0.90}\\ 
\bottomrule
\end{tabular*}
\end{table}
where $\alpha_\ast^{GT}$ {represents} the posture along the *-axis measured with the motion capture, $\alpha_\ast^e$ is the posture reconstructed by using the encoders and neglecting any secondary interaction, $p_{\ast i}$ are the identified linear coefficients that fit the data, and $\varepsilon_\ast$ is the residual of the fit.
Table~\ref{tab:posture_reconstruction} {and Figure~\ref{fig:PostureReconstruction}} report the output of the estimation. Note that $R^2$ is close to 1, {indicating a significant fit}. Moreover, the bias coefficients $p_{\ast 0}$ are close to 0, and the direct proportional coefficients $p_{yy}$ and $p_{zz}$ are close to 1, {suggesting} that $\alpha_\ast^e$ is accurate. 
Still, we could improve the estimate by exploiting the additional information about the dependency of the pose from the stiffness and the second DoF.

\section{Experimental Validation}\label{sect:experimental_validation}
To validate the control strategy, we {replicated} the experiments {detailed} in Section~\ref{sect:calibration} after applying the calibration procedure.

\begin{figure}[!t]
    \centering
    \subfloat[]{\label{fig:sine_comp}\includegraphics[width = 0.45 \linewidth]{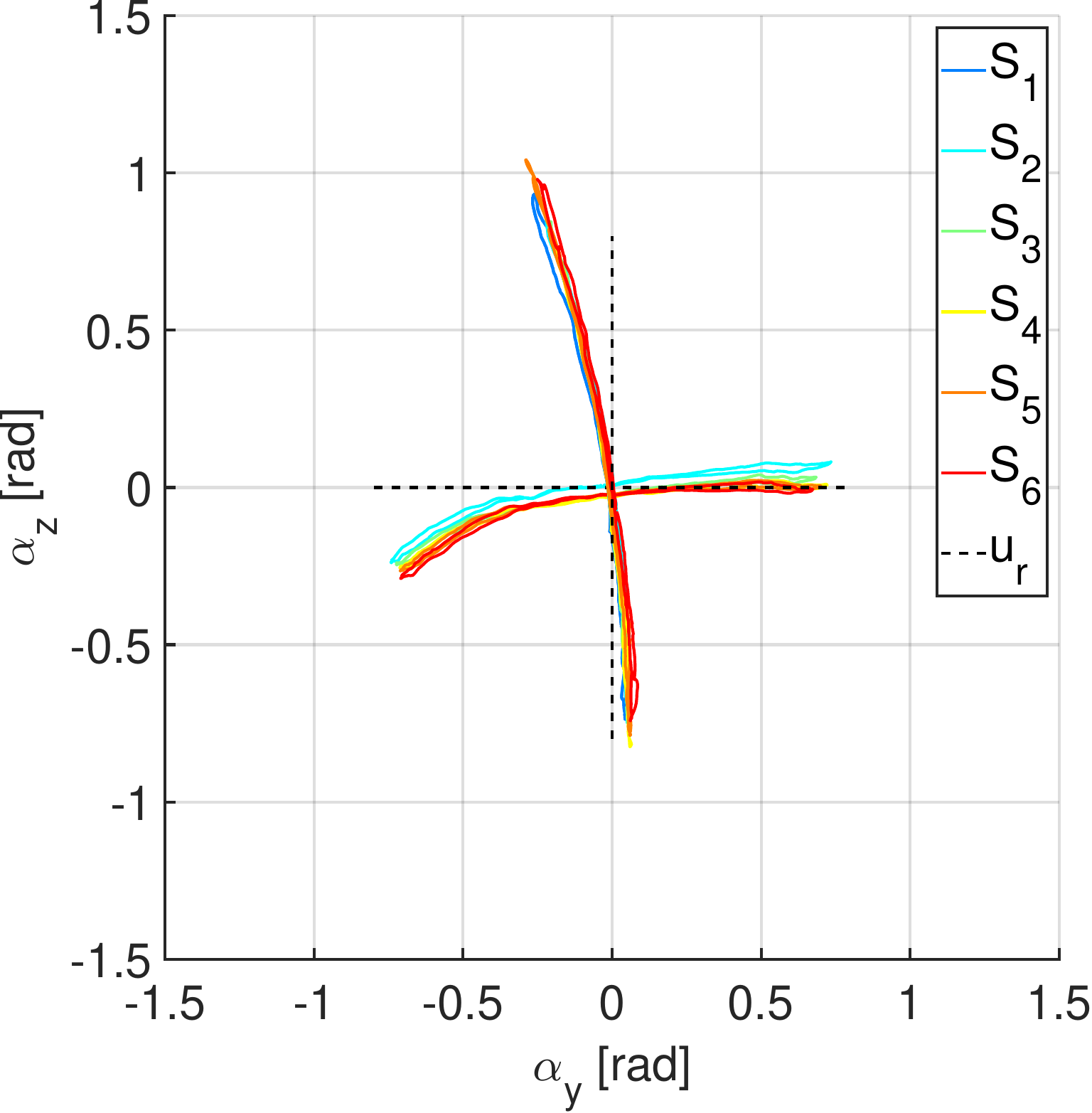}}
    \hfil
    \subfloat[]{\label{fig:GridComp}\includegraphics[width = 0.45 \linewidth]{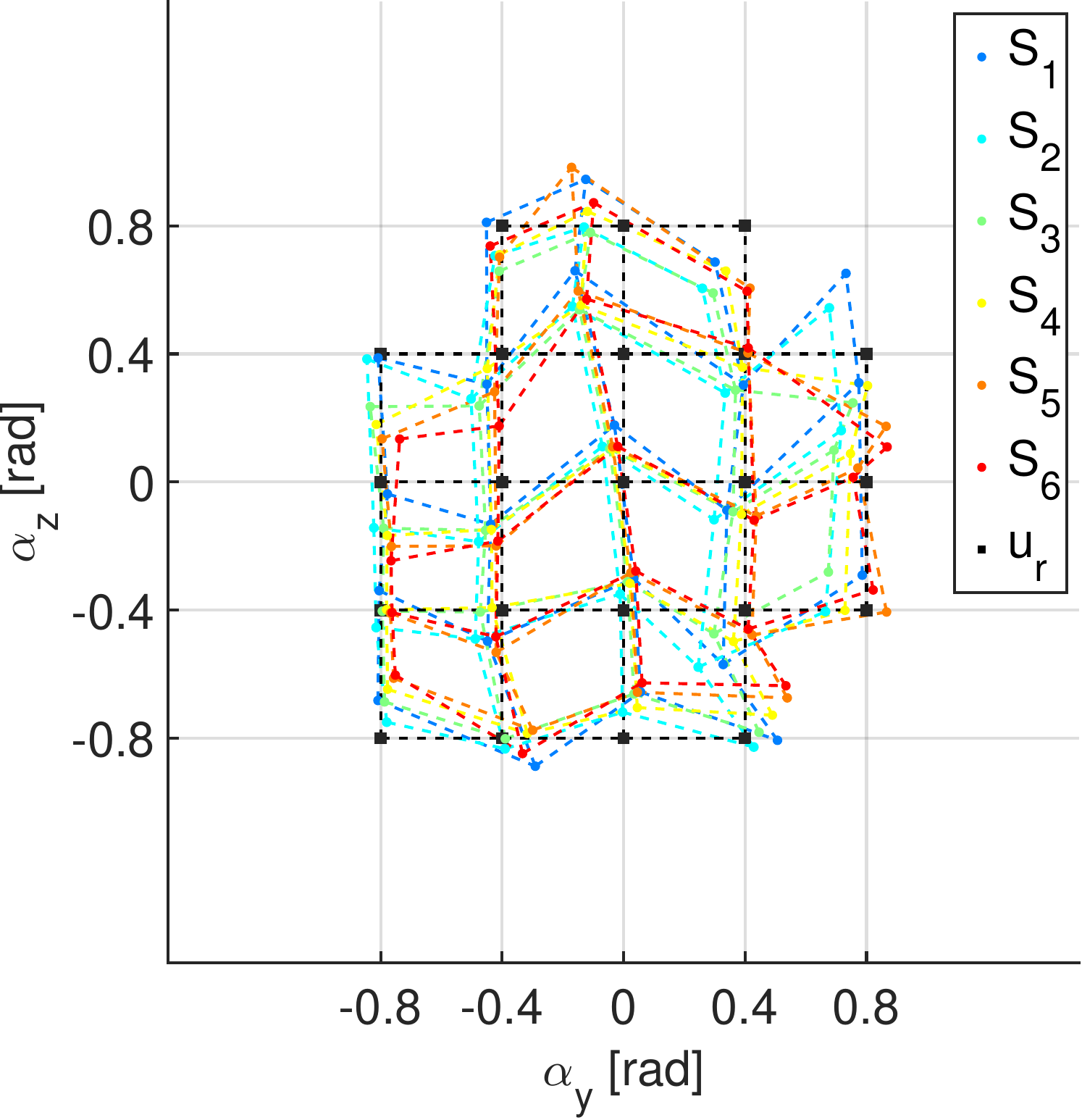}}
    \caption{Characterization of the device kinematic behavior after system calibration. Each color corresponds to a distinct stiffness configuration (increasing from $S_1$ to $S_6$). Panel (a) shows the device chasing sinusoidal references, acting on a single axis at a time. Panel (b) shows the experimental RoM of the prototype, acquired by sampling its 2D workspace with a grid.}
    \label{fig:posture_control_Comp}
\end{figure}

Figure~\ref{fig:sine_comp} reports the posture of the device while chasing sinusoidal references acting, one per time, on both DoFs. To {assess} the accuracy of the posture pursuit, we applied the Linear Fit Method (LFM) as in \cite{iosa2014assessment}. Precisely, we fit the posture reference with a linear model of the effective posture. {Consequently}, if the tracking is accurate, the linear coefficient will be close to 1 and the bias close to 0. The results of the linear fit, resumed in the upper half of Table~\ref{tab:system_val}, confirmed that the posture of the device is an accurate replica of the desired pose.
{Additionally}, we applied the LFM to compare the posture estimated using the encoders with the measurements of the motion capture system. The results, {presented} on the right side of Table~\ref{tab:system_val}, confirmed that the posture is reconstructed accurately.

Figure~\ref{fig:GridComp} shows the performance of the VS joint tracking simultaneous references on both axes. Similarly, we applied the LFM to evaluate the accuracy of posture tracking and estimation. The identified parameters, reported in the lower half of Table~\ref{tab:system_val}, {affirmed the effectiveness} of the calibration procedure, validating both posture control and reconstruction strategies.
{Notably, the experimental data did not reveal any correlation between the commanded stiffness and posture tracking and reconstruction performance or the workspace.}

The PS unit achieves very precise ($\sigma_{PS} = 0.01 $ rad) and accurate control ($RMSE(\theta_{PS})=0.02$ rad), and its RoM is unconstrained ($\pm\pi$).

The output speed of the device is computed using step references that {span} the RoM of each individual DoF at minimum and maximum stiffness configurations. The {wrist speed} decreases as the stiffness {increases}, except for the PS, {as} its mechanical transmission is decoupled from the VS units. The average output speed ranges from 4.0 to 8.4 rad/s for the FE and from 3.0 to 6.3 rad/s for the RUD, while the PS average speed is 10.8 rad/s.

\begin{table*}[!t]
\small\sf\centering
\caption{Validation of posture control and reconstruction using the LFM. $a_\ast$ denotes the linear parameter of the LFM and $b_\ast$ the bias. Data for both sinusoidal and grid references are presented separately for both posture control and reconstruction. The standard errors of the identified coefficients are reported within the parentheses.}
\label{tab:system_val}
\begin{tabular*}{\textwidth}{@{\extracolsep{\fill}} c | c c c c | c c c c}
\toprule
 & \multicolumn{4}{c|}{Posture Control}  & \multicolumn{4}{c}{Posture Reconstruction} \\
\midrule
Parameter & $a_{\ast}$ & $b_{\ast}$& $RMSE$ & $R^2$ & $a_{\ast}$ & $b_{\ast}$& $RMSE$ & $R^2$ \\
\midrule
 & \multicolumn{8}{c}{Sinusoidal Reference} \\
\midrule
$\alpha_y$ & 1.01 (1.4e-3) & 0.04 (5.0e-3) & 0.08 & 0.95 & {1.01 (1.4e-3)}& {0.05 (4.8e-4)} & {0.08} & {0.95} \\
$\alpha_z$ & 0.88 (1.3e-2) & 0.01 (5.0e-3) & 0.08 & 0.94 &{0.84 (1.6e-3)} & {-0.05 (6.6e-4)} & {0.11} & {0.91} \\
\midrule
 & \multicolumn{8}{c}{Grid Reference}\\
\midrule
$\alpha_y$ & 0.98 (0.01) & -0.02 (5.7e-3) & 0.06 & 0.98 & 1.03 (0.02) & 3.7e-3 (8.2e-3) & 0.09 & 0.97 \\
$\alpha_z$ & 0.92 (0.02) & -0.02 (0.01) & 0.12 & 0.940 & 0.94 (0.03) & -4.0e-3 (0.02) & 0.18 & 0.88\\
\bottomrule
\end{tabular*}
\end{table*}

\subsection{Variable Stiffness Ability}\label{sect:variablestiffnessability}
\begin{figure}[!t]
    \centering
    \includegraphics[width = 0.95 \linewidth]{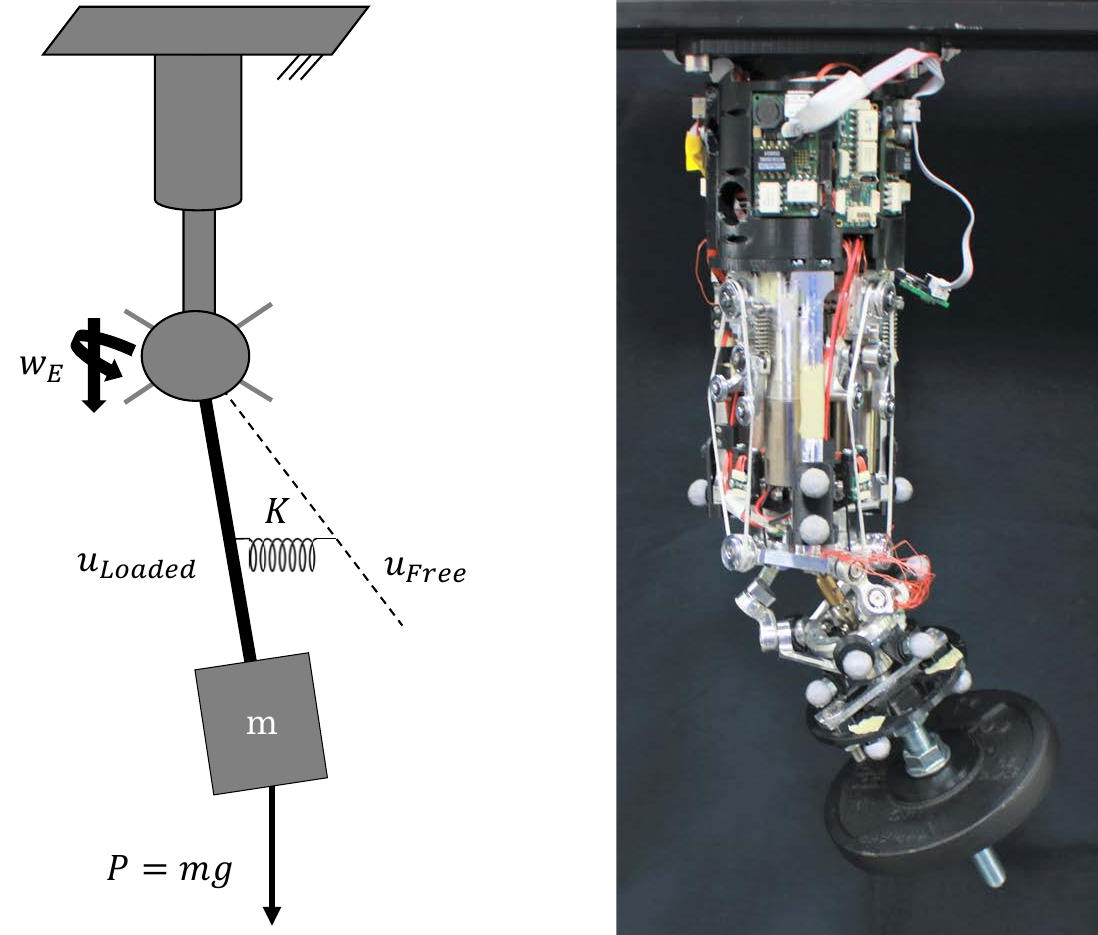}
    \caption{Experimental setup employed to quantify the variable stiffness ability of the device. The VS-Wrist is suspended upside-down from a fixed frame, with a mass attached to the coupler. The external load causes the coupler to shift its equilibrium configuration coherently with the commanded stiffness, exploiting the embedded elasticity.}
    \label{fig:StiffnessExperimentalSetup}
\end{figure}

\begin{figure}[!t]
    \centering    
    \subfloat[]{\label{fig:GridLoaded_v2} \includegraphics[width = 0.48 \linewidth]{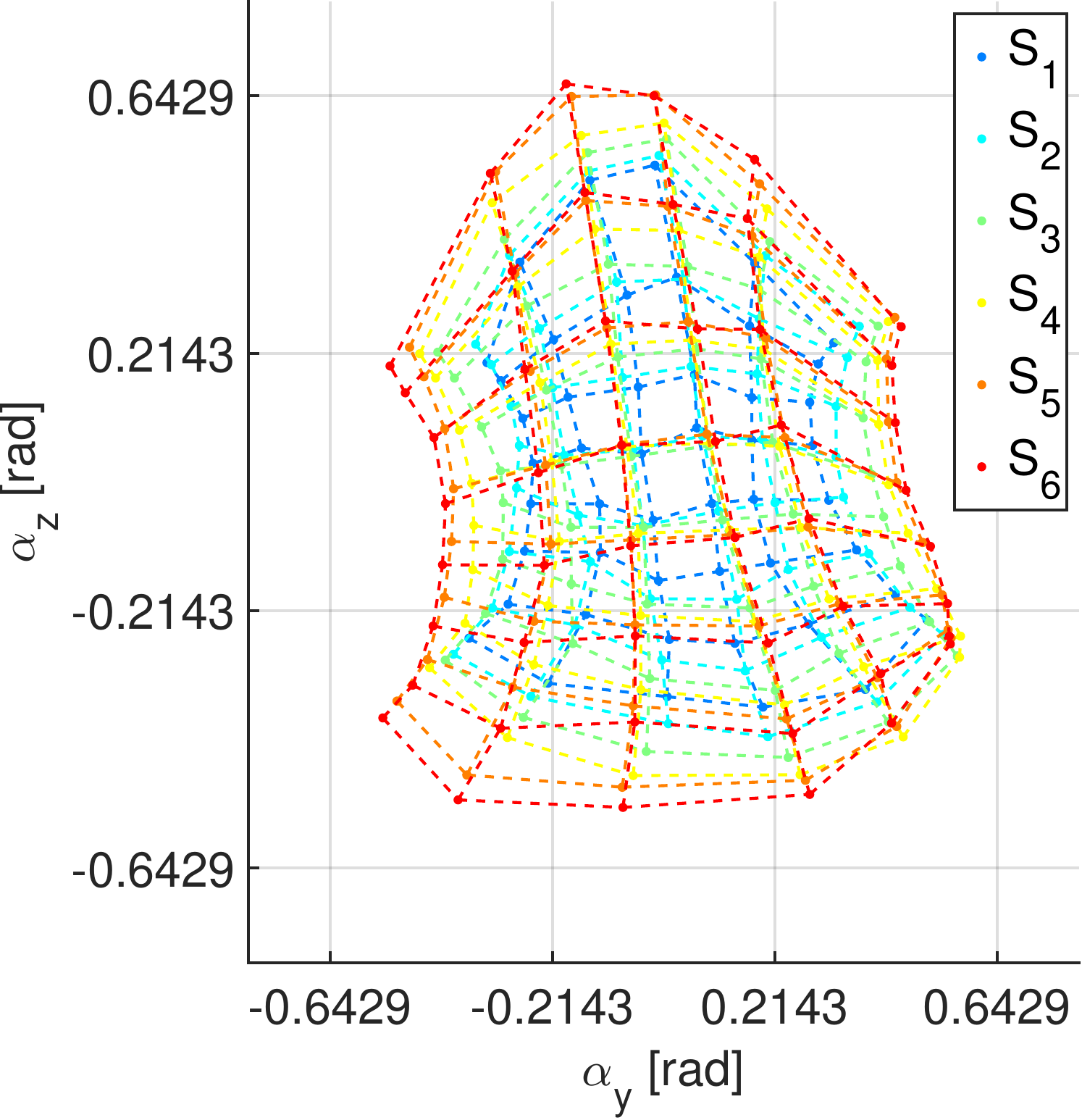}} 
    \hfil 
    \subfloat[]{\label{fig:StiffCircles}\includegraphics[width = 0.48 \linewidth]{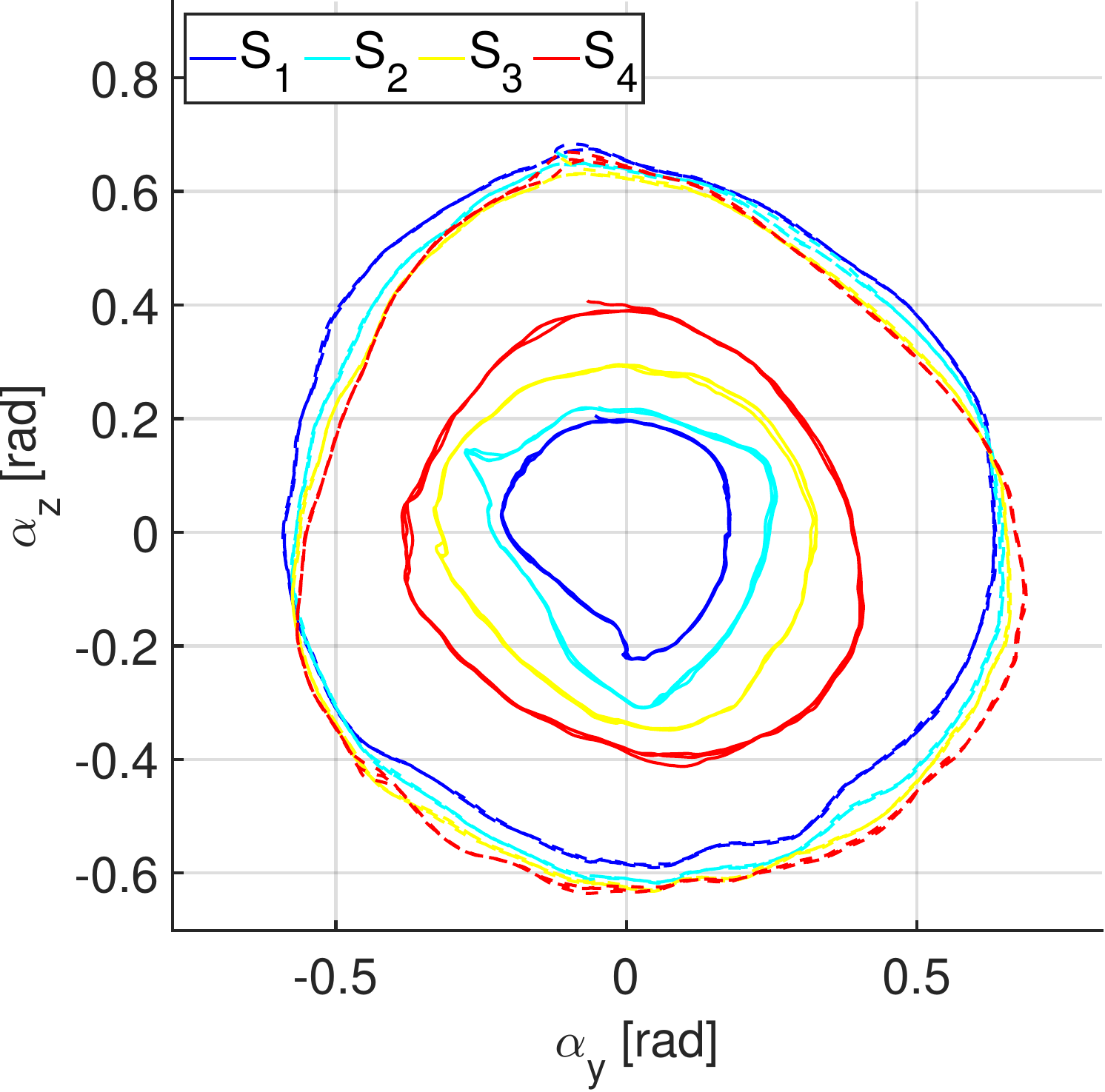}}
    \caption{ Experimental evaluation of the device variable stiffness ability. Panel (a) reports the posture of the VS joint as it explores its 2D-RoM at various stiffness configurations (increasing from $S_1$ to $S_6$) while a load is attached to the end-effector. Panel (b) shows the device tracking circular trajectories at four different stiffness levels (increasing from $S_1$ to $S_4$) in both unloaded (dashed lines) and loaded (continuous lines) configurations. The radii of the latter circles are proportional to the stiffness, which defines the resistance to external perturbations.}
    \label{fig:Stiffness_Exp}
\end{figure}

\begin{table}[!t]
\small\sf\centering
\caption{Stiffness range of the device, statistic significance of the fit, and comparison with human wrist stiffness (extracted from \cite{pando2014position}). The stiffness matrix coefficients are reported in Nm/rad and the $RMSE(\Delta w)$ in Nm. Results are reported for the minimum and maximum stiffness configurations, denoted as $S_m$ and $S_M$. The standard deviation of the estimated coefficients is indicated within the parentheses.}\label{tab:Stiffness}
\begin{tabular}{c | c c | c}
\toprule
 Parameter & $S_{m}$ & $S_{M}$ & Human\\
\midrule
$K_{yy}$ & 0.27 (0.01) & 0.93 (0.04) &  2.02 (0.19) \\
$K_{yz} = K_{zy}$ & -0.35 (0.02) & -0.51 (0.05)  & -0.27  (0.04)\\
$K_{zz}$ & 0.24 (0.01) & 0.72 (0.04) & 0.85 (0.07)\\
$RMSE(\Delta w)$ & 0.03 & 0.05 & \\
$R^2$ & 0.94 & 0.91 & 0.98 \\
\bottomrule
\end{tabular}
\end{table}

The VS-Wrist exploits {its} redundant elastic actuation to vary the stiffness of the coupler, adapting to tasks of different natures. {To demonstrate this capability}, we {replicated} the experiment {depicted} in Figure~\ref{fig:GridUnloaded} after attaching a load to the end-effector. During this experiment, the VS-Wrist is hung upside-down on a fixed frame, as shown in Figure~\ref{fig:StiffnessExperimentalSetup}. An external load of 640~g is {affixed} to the coupler at a distance of 78~mm.
Therefore, the force of gravity applies a known external wrench that varies with the posture of the wrist.
To quantify the stiffness, we assume that it remains constant during each trial. 
This assumption holds
\begin{equation} \label{eq:StiffExp}
    \Delta w_u = K_u \Delta u \enspace,
\end{equation}
where $\Delta w_u, \Delta u \in \mathbb{R}^{2\times N_{sample}}$ are matrices whose columns contain, for each sample of the trial, the difference between the loaded and unloaded condition of {the external wrench and posture, respectively,} along the minimum parametrization $u$, and $K_u$ represents the angular stiffness of the device. 
Therefore, inverting \eqref{eq:StiffExp} yields the stiffness matrix $K_u$ that best fits the experimental data. 
Table~\ref{tab:Stiffness} reports the range of joint stiffness and the statistical significance of its estimation.

Figure~\ref{fig:StiffCircles} shows the device tracking circular trajectories {under} both loaded and unloaded conditions at various stiffness configurations.
The device stiffness dictates its response to external perturbations, with higher stiffness resulting in smaller differences between trajectories in loaded and unloaded configurations.
{Accordingly}, in Figure~\ref{fig:StiffCircles}, the circles reduce their size inversely proportional to the commanded stiffness when the load is applied.

\subsection{Qualitative evaluation of Motion and Stiffness Behavior}
\begin{figure*}
    \centering
    \includegraphics[width = 0.98 \linewidth]{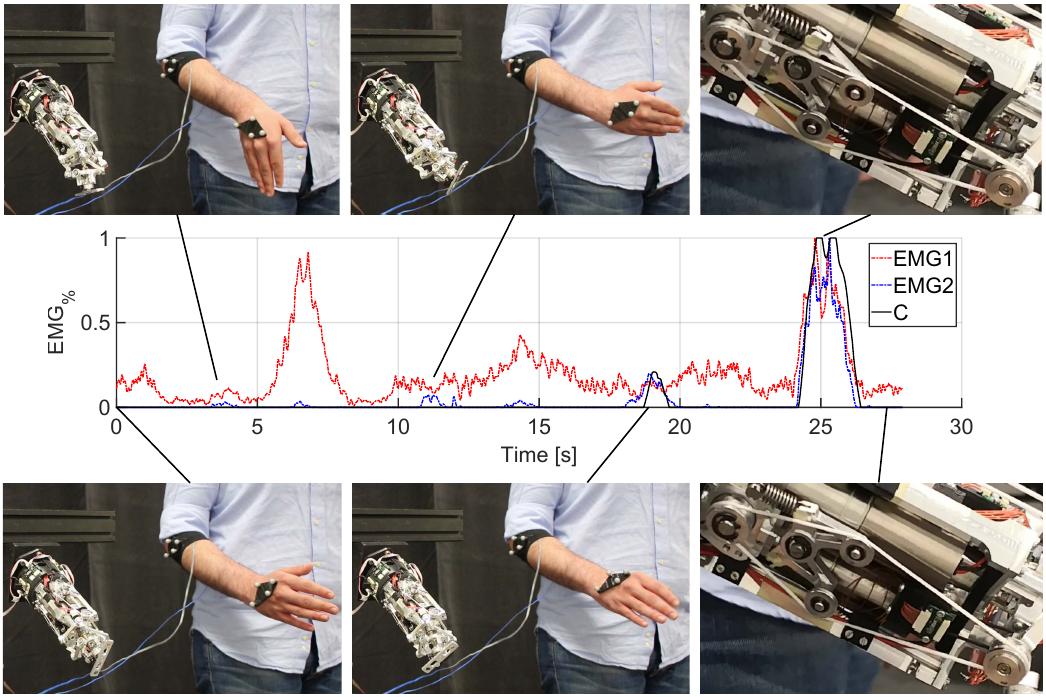}
    \caption{Photosequence capturing the VS-Wrist tracking the user's posture and muscular cocontraction. The central graph displays the corresponding normalized EMG signals measured from a couple of antagonistic muscles of the user's forearm, highlighting the muscular cocontraction C, which defines the stiffness reference of the robotic wrist. Each panel depicts a specific instant of the experiment and is associated with the corresponding EMG measurement. The panels on the right describe the functioning of the elastic transmission that achieves stiffness regulation. The bottom-right frame shows the device in the soft configuration, where tendon tension is low. The top-right panel portrays the VS-Wrist in the rigid configuration, matching the operator's muscular impedance. Coherently, the tendons on one side of all VS units stretch to increase the stiffness.}
    \label{fig:Sequence_Teleimp_NoHandEMG}
\end{figure*}

\begin{figure*}
    \centering
    \subfloat[]{\includegraphics[width = 0.32 \linewidth]{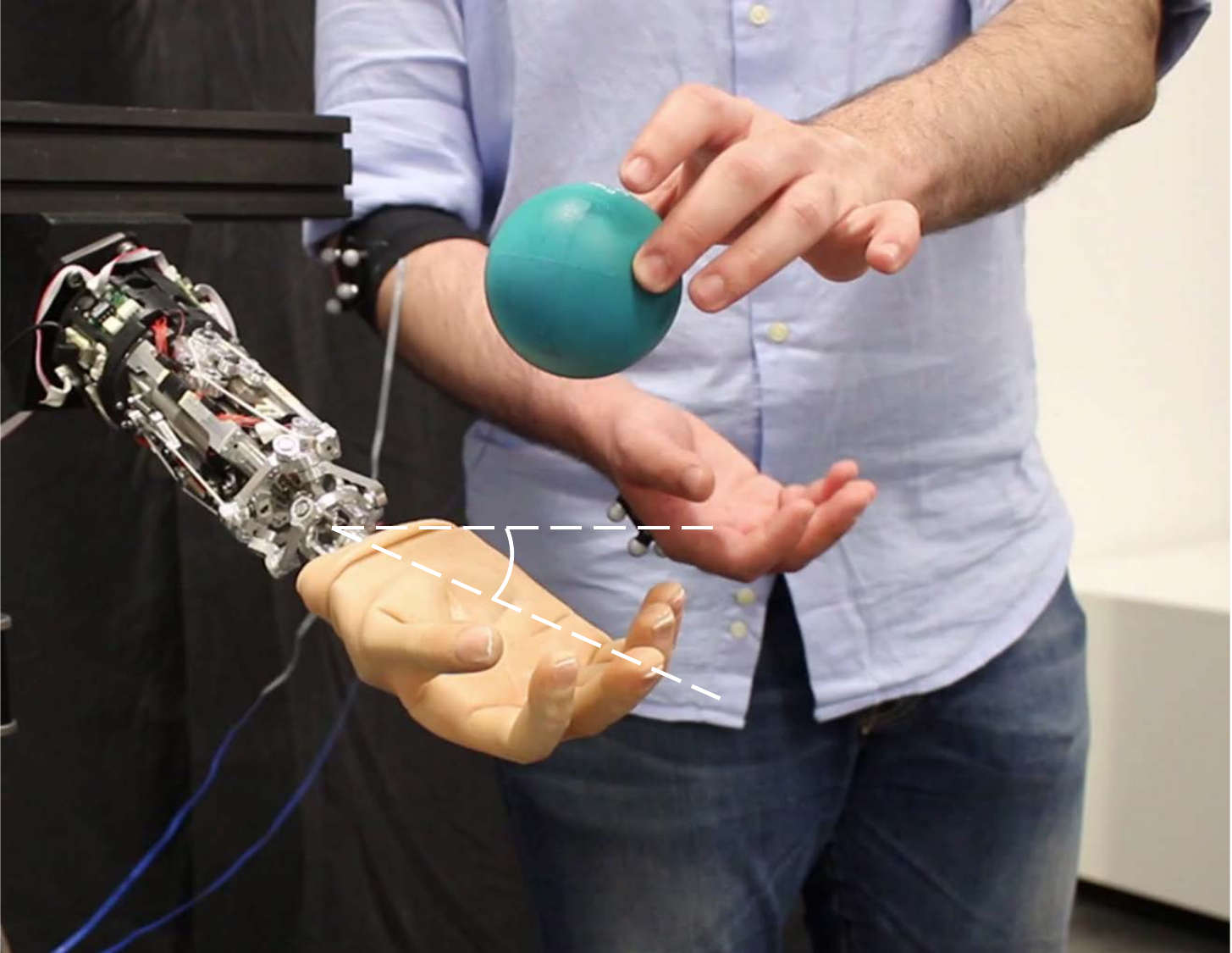}} \hfil
    \subfloat[]{\label{fig:Sequence_VSBallVideo_Soft}\includegraphics[width = 0.32 \linewidth]{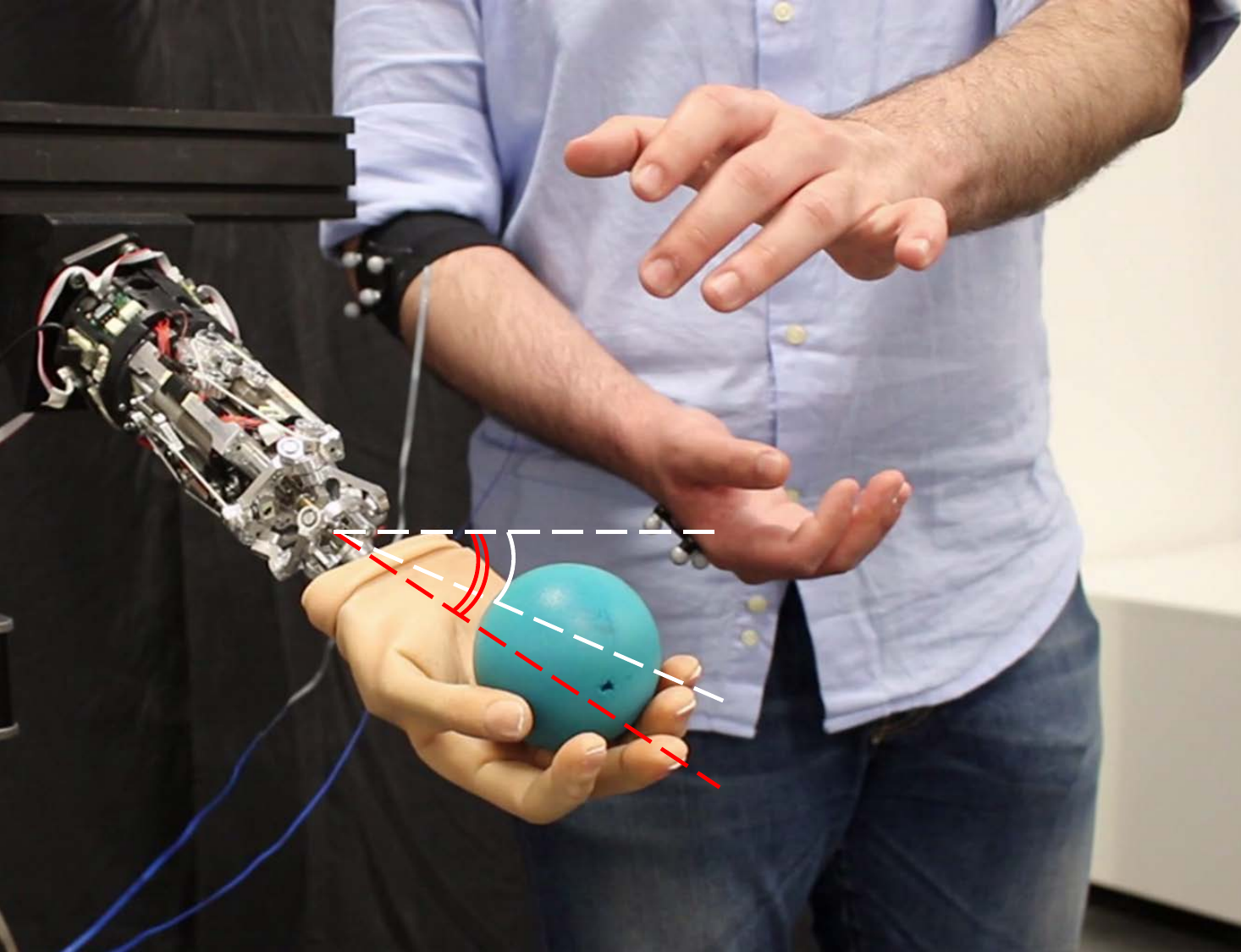}} \hfil
    \subfloat[]{\label{fig:Sequence_VSBallVideo_Rigid}\includegraphics[width = 0.32 \linewidth]{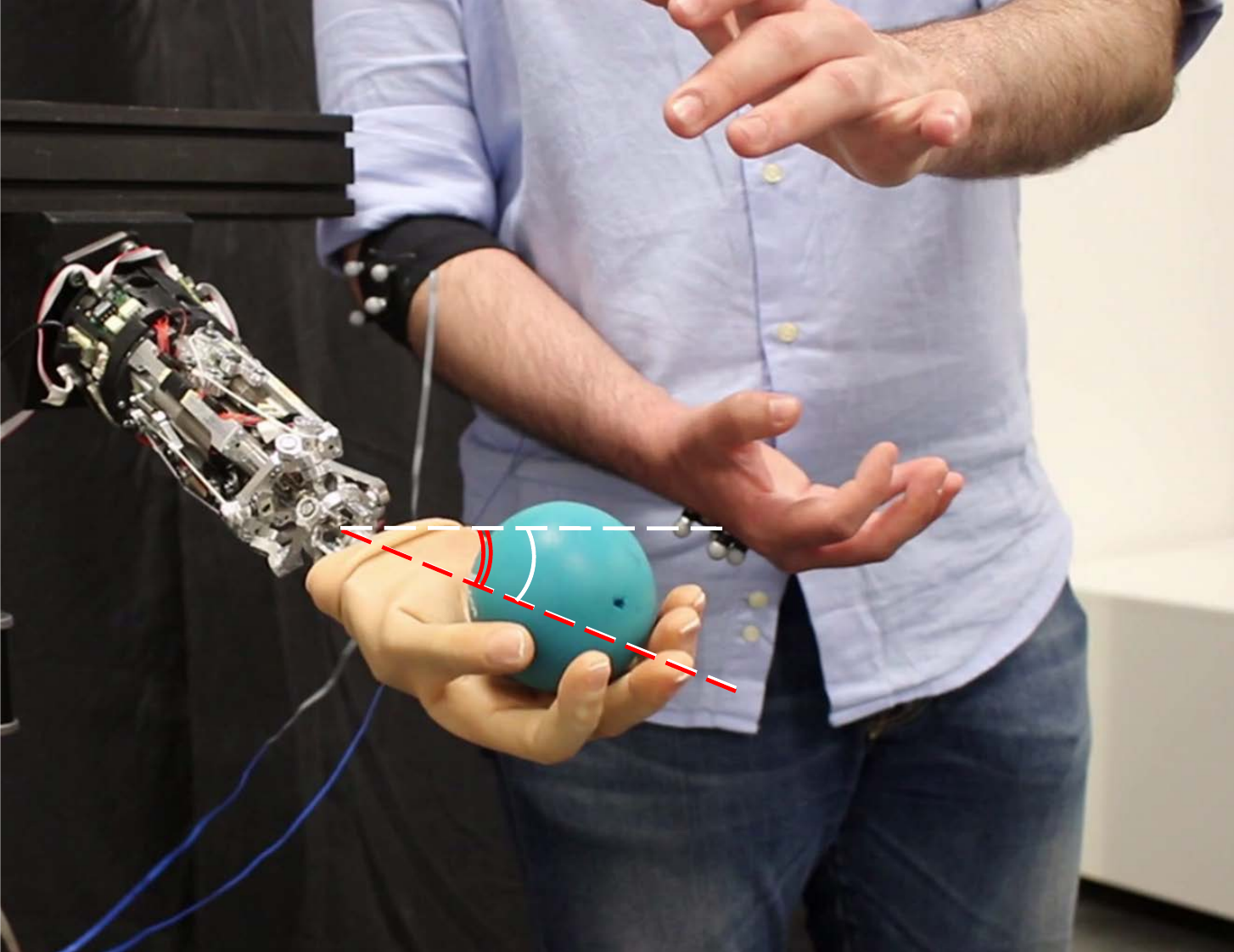}}     
    \caption{Response to external perturbation at different stiffness configurations. Visual feedback is provided by the angle formed between the segment from the center of the coupler to the middle finger and the horizontal plane. During the trial, the VS-Wrist holds a reference posture (panel (a), white lines) before being impacted by a ball in free fall. Due to variable stiffness, the device reaches different equilibrium configurations (red lines), as illustrated in panel (b) for the soft configuration and (c) for the rigid one.}
    \label{fig:Sequence_VSBallVideo}
\end{figure*}

To provide a more intuitive representation of the device operating principle and emphasize its anthropomorphism, {we map} the posture and stiffness of a human operator's wrist into command signals for the robotic device. The reference posture is captured with cameras through a marker set attached to the user's forearm and hand.
Furthermore, we registered the operator's myoelectric activity of the wrist flexor and extensor to measure {their} cocontraction, which we related to the desired stiffness of the VS-Wrist, similarly to \cite{capsi2020exploring}. To do so, we employed two commercial electrodes 13E200 from Ottobock, which are commonly used to control prosthetic limbs. 
Figures~\ref{fig:Sequence_Teleimp_NoHandEMG} and \ref{fig:Sequence_VSBallVideo} {summarize} the outcome of the experiments with some frames of the video attached as supplemental material to this paper. 
Figure~\ref{fig:Sequence_Teleimp_NoHandEMG} shows that the spring preload is modulated through muscular cocontraction while the device tracks the operator's wrist posture. Figure~\ref{fig:Sequence_VSBallVideo} proves that different stiffness configurations achieve various responses to external perturbations. Therefore, the variable stiffness should enhance task adaptability and environmental interactions.


\section{Discussion}\label{sect:discuss}
{The initial experimental phase revealed non-idealities,} that we addressed and mitigated through a calibration process. The calibration of the stiffness controller, {compensating for VS unit asymmetry}, successfully eliminated {undesired} movements associated with changes in joint stiffness. The combined control calibration enhanced {posture accuracy} by leveraging the {experimental} motion behavior, identified {through grid sampling of the RoM.}
Undoubtedly, the absence of posture feedback within the control law {induces} a residual error in the end-effector trajectory. Conversely, our decision to prioritize a simple implementation not only maintains the intrinsic compliance of the device but also demands low computational power, making it well-suited for prosthetic applications. Moreover, the inaccuracy in tracking a desired posture becomes a minor concern in applications where a human operator closes the control loop, such as prosthetics or teleoperation.

The experimental evaluation confirmed that the VS-Wrist can achieve 3 DoFs in position while continuously changing {its} overall stiffness. Accurate real-time posture reconstruction was achieved using measurements of the first joint angles and forward kinematics.
{Furthermore, Figure~\ref{fig:Sequence_Teleimp_NoHandEMG} highlights that the VS-Wrist} functioning resembles human motion and impedance regulation principles. 
Although the {concept} of adaptively controllable impedance in prostheses dates back to the early eighties \cite{hogan1983prostheses}, its practical implementation has been hampered by technological limitations. Table~\ref{tab:comp_wrists} compares the features of the VS-Wrist with those of the human and other 3 DoFs prosthetic wrists. The theoretical output torque is sufficient to manipulate a load of 1.8 kg with nominal motor torques and of 5.2 kg with maximum motor torques. Note that, due to the non-backdrivability of the actuation system, active motor torque is required only to move from the equilibrium condition, while static loads are counteracted by the friction-based mechanism. The VS-Wrist speed and RoM align with the functional needs of the human wrist. Additionally, its size and weight are smaller than those of the human forearm and are comparable to other 3 DoFs architectures documented in the literature. Despite these similarities, our first prototype presents limitations hindering its prosthetic application.

\begin{table*}[!t]
\small\sf\centering
\caption{Comparative table showcasing human and 3 DoFs prosthetic wrists. Apart from the VS-Wrist, all the other devices feature rigid and fixed impedance actuation. This table has been adapted from \cite{damerla2022design}. The tested RoM of the prosthetic devices is reported inside square brackets. Cells are left empty when information is missing at the source.}
\label{tab:comp_wrists}
\begin{tabular*}{\textwidth}{@{\extracolsep{\fill}} c c c c c c c c c c}
\toprule
Name & DoFs & RoM (°) & D (mm) & L (mm) & m (g) & $\tau$ (Nm)& $\omega$ (rad/s) \\
\midrule
\begin{tabular}{@{}c@{}}Median Male \\Wrist and Forearm \end{tabular} & \begin{tabular}{@{}c@{}} PS \\ FE \\ RUD \\\end{tabular} & \begin{tabular}{@{}c@{}} 83/100 (61/75)${}^\ast$ \\ 76/73 (54/48)${}^\ast$  \\ 25/45 (22/38)${}^\ast$  \\\end{tabular}  & 98 & 269 & 1420 & \begin{tabular}{@{}c@{}} 9/9.5 \\ 12.7/7.9  \\ 13.0/12.4   \\\end{tabular} & 
\begin{tabular}{@{}c@{}} 38/33 \\ 27/26 (1.7)${}^\ast$  \\ 10/11 (1.7)${}^\ast$   \\\end{tabular} \\
\midrule
\begin{tabular}{@{}c@{}}Median Female \\Wrist and Forearm \end{tabular} & \begin{tabular}{@{}c@{}} PS \\ FE \\ RUD \\\end{tabular} & \begin{tabular}{@{}c@{}} 83/100 (61/75)${}^\ast$ \\ 76/73 (54/48)${}^\ast$  \\ 25/45 (22/38)${}^\ast$  \\\end{tabular}  & 86 & 243 & 1060 & \begin{tabular}{@{}c@{}} 4.5/4.6 \\ 8.8/5.8  \\ 8.2/8.0   \\\end{tabular} & 
\begin{tabular}{@{}c@{}} 38/33 \\ 27/26 (1.7)${}^\ast$  \\ 10/11 (1.7)${}^\ast$   \\\end{tabular} \\
\midrule
\begin{tabular}{@{}c@{}}VS-Wrist \\ \end{tabular} & \begin{tabular}{@{}c@{}} PS \\ FE \\ RUD \\\end{tabular} & \begin{tabular}{@{}c@{}} 360+  \\ 90/90 [55/45]  \\ 90/90 [48/48]  \\\end{tabular}  & 70 & 170 & 1110 & \begin{tabular}{@{}c@{}} 2 \\ 2.4   \\ 2.7  \\\end{tabular} & 
\begin{tabular}{@{}c@{}} 10.8 \\  4-8.4  \\ 3-6.3   \\\end{tabular} \\
\midrule
\begin{tabular}{@{}c@{}} \cite{bajaj2018design}  \end{tabular} & \begin{tabular}{@{}c@{}} PS \\ FE \\ RUD \\\end{tabular} & \begin{tabular}{@{}c@{}} 360+  \\ 90/90 [$\geq$ 40/40] \\ 90/90 [$\geq$ 40/40]  \\\end{tabular}  & 86 & 180 & 578 & \begin{tabular}{@{}c@{}} \\ \\   \\\end{tabular} & 
\begin{tabular}{@{}c@{}} 1.84 \\  $\geq$ 0.6  \\ $\geq$ 0.6    \\\end{tabular} \\
\midrule
\begin{tabular}{@{}c@{}} \cite{abayasiri2017mobio}\\   \end{tabular} & \begin{tabular}{@{}c@{}} PS \\ FE \\ RUD \\\end{tabular} & \begin{tabular}{@{}c@{}} [70/85]  \\ 
 \text{[60/60]} \\ \text{[25/27]}  \\\end{tabular}  &  & 182  & \begin{tabular}{@{}c@{}} 3200  \\ Total for Hand,\\ Wrist, Elbow  \\\end{tabular}  & & \\
\midrule
\begin{tabular}{@{}c@{}} \cite{fite2008gas}\\   \end{tabular} & \begin{tabular}{@{}c@{}} PS \\ FE \\ RUD \\\end{tabular} & \begin{tabular}{@{}c@{}} 150  \\ 170 \\ 60  \\\end{tabular}  & &  & \begin{tabular}{@{}c@{}}  2000 \\ Total for Hand, \\ Wrist, Elbow  \\\end{tabular}  & \begin{tabular}{@{}c@{}}  4.2 \\ 4.5 \\ 4.5  \\\end{tabular} \\
\bottomrule
\multicolumn{8}{l}{\begin{tabular}{@{}l@{}}  \footnotesize{${}^\ast$ Functional for standard activities of daily living.} \end{tabular}}\\
\end{tabular*}
\end{table*}

\subsection{Prototype Limitations}\label{sect:prot_limit}
We expect several mechanical refinements for the next release of the prototype. The current version exhibited limitations in both maximum payload and stiffness, primarily attributed to high manufacturing tolerances {resulting in} mechanical backlashes.
For this reason, attempts to replicate experiments detailed in Section \ref{sect:variablestiffnessability} with the wrist positioned horizontally were unsuccessful. The device could only sustain the payload at high stiffness levels and within a limited RoM, precluding the identification of its stiffness.
Additionally, despite incorporating the functional RoM of the human wrist, the prototype workspace and accuracy were reduced due to non-nominal kinematics.

Although the VS-Wrist approximately matches half the size and weight of a human forearm, future designs should optimize its weight for enhanced user comfort. Furthermore, the prototype length may pose challenges {in} fitting most transradial amputees but could be suitable for transhumeral amputees.

As this paper presents our first prototype, certain aspects require refinement in both design and manufacturing before undergoing clinical trials. Nonetheless, with proper optimization, the proposed architecture holds promise for prosthetic applications, introducing innovative features such as variable stiffness and 3 DoFs movements. This makes it a significant contribution to the field of prosthetics{, as} the presented design has the potential to considerably enhance the dexterity and naturalness of upper limb prostheses, addressing the limitations of current commercial devices that predominantly feature passive, rigid joints with restricted DoFs.

\section{Conclusion and Future Developments}\label{sect:conclusion}
This study addresses the challenge of modeling and controlling a variable stiffness 3 DoFs wrist featuring redundant elastic actuation. Its controllable stiffness and inherent compliance enable {the device to interact safely} with the environment, adapting to tasks of various natures. In the soft configuration, contact forces are minimized, while the device effectively resists external perturbations at high stiffness. The presented calibration procedure successfully compensated for the prototype deviations from its nominal model, ensuring precise posture control and reconstruction.

Optimizing the design and manufacturing processes is crucial to reduce mechanical backlash and augment the stiffness of the device. Attention must be directed towards minimizing the device weight, considering the potential discomfort for prosthesis users. Additional investigations into the robustness of the device are necessary to ensure proper functionality during extended use. 
Nevertheless, the VS-Wrist {holds} the potential to enhance the natural feel of existing bionic limbs by resembling the kinematic and compliant behaviors of its natural counterpart. Further quantitative analyses involving standard daily activities could provide valuable insights into assessing the morphological and functional similarities between the device and the human wrist.

Other developments of this work concern the mapping of user EMG signals into control commands for the prosthesis. While coordinated control of multiple DoFs remains an open challenge, recent advancements in neuroscience offer potential {solutions} for operating more complex prostheses.

\begin{acks}
The authors would like to thank Vinicio Tincani, Mattia Poggiani, Manuel Barbarossa, Emanuele Sessa, Marina Gnocco, Giovanni Rosato, and Anna Pace for their fundamental technical support given during the hardware implementation of this work. An additional thanks goes to Simon Lemerle for his work on the design concept of the device.
\end{acks}

\begin{dci}
The authors declare that the research was conducted in the absence of any commercial or financial relationships that could be construed as a potential conflict of interest.
\end{dci}

\begin{funding}
This research has received funding from the European Union’s ERC program under the Grant Agreement No. 810346 (Natural Bionics). The content of this publication is the sole responsibility of the authors. The European Commission or its services cannot be held responsible for any use that may be made of the information it contains.
\end{funding}

\begin{sm}
The Supplementary Material for this article, containing a video demonstration of the prototype and mathematical details about the static equilibrium equations and the elastic transmission model, can be downloaded from the dedicated section of this website.
\end{sm}

\theendnotes

\bibliographystyle{SageH}
\bibliography{References.bib}
 \end{document}


\runninghead{Milazzo et al.}
\onecolumn
\appendix
\section*{Appendix A: Derivation of the Static Equilibrium Equations} \label{app: Appendix A}
This appendix details the derivation of the static equilibrium equations reported in Section 3.5. To start, we compute the Jacobian matrix $J_P(q)$ of each leg of the manipulator in the coupling point $P$ with the end-effector. These matrices solely depend on the joint variables of the respective leg and relate their joint velocities to the twist of the coupling point. Exploiting the kinetostatic duality, their transpose relates the torques acting on the joints $\tau$ with the wrench $w_P$ acting on the coupling point, such that $\tau = J_P^\top(q) w_P$. Employing the Denavit Hartenberg convention, for a revolute joint, the ith column of the Jacobian matrix can be computed as
\begin{equation}
\label{eqn:DH_Jacob}
    J_P^i = 
    \begin{bmatrix}
    k_{i-1} \times O_{i-1}P\\
    k_{i-1}
    \end{bmatrix} \enspace,
\end{equation}
where $k_{i-1}$ represents the versor of $Z_{i-1}$-axis and $O_{i-1}P$ is the distance of the considered point $P$ from the origin of frame $\{S_{i-1}\}$. All geometrical quantities are expressed in the fixed base frame $\{S_b\}$.
The position of the coupling point $P$ w.r.t. the local frame $\{S_{3}\}$ remains constant and is equal to
\begin{equation}
    O_0P = \text{T}_b^3 (q) \begin{bmatrix}
       0\\
       0\\
       r_c\\
       1
    \end{bmatrix} \enspace,
\end{equation}
where  $r_e = 22.5$ mm is the coupler radius. Similarly, for the subsequent local frames it holds
\begin{equation}
    O_iP = O_0P - \text{T}_b^i(:,4) \quad \quad \text{for} \enspace  i = 1,2,3 \enspace,
\end{equation}
where $\text{T}_b^i(:,4)$ denotes the fourth column of the transformation matrix, hence the position of $O_i$ w.r.t. the reference frame $\{S_b\}$.
Accounting for the equilibrium of the entire manipulator yields
\begin{equation}
\label{eqn:joint_jac}
\tau = 
    \begin{bmatrix}
    \tau_A \\
    \tau_B \\
    \tau_C
    \end{bmatrix}
    = 
    \begin{bmatrix}
    J_P^{A}(q_A)^\top &  0_{6\times4} &  0_{6\times4} \\
    0_{6\times4} & J_P^{B}(q_B)^\top &   0_{6\times4} \\
    0_{6\times4} &   0_{6\times4} &   J_P^{C}(q_C)^\top
    \end{bmatrix}
    \begin{bmatrix}
    w_A \\
    w_B \\
    w_C
    \end{bmatrix}
    = J^\top (q) w\;,
\end{equation}
where $\tau_* \in \mathbb{R}^{4}$ represent the joint torques acting on the leg *, $w_* \in \mathbb{R}^{6}$ is the wrench acting on the coupling point of the same leg, and $J^\top(q) \in \mathbb{R}^{12\times12}$ is the complete Jacobian of the parallel manipulator.
We introduce the matrix $G_*$ which conveys the effect of the wrench $w_*$ acting on $P_*$, to the center of the end-effector $O_e$ as 
\begin{equation}
    G_* = 
    \begin{bmatrix}
    I_{3\times3} & 0_{3\times3} \\
    \widehat{O_eP_*} & I_{3\times3}
    \end{bmatrix} \enspace,
\end{equation}
where $\widehat{O_eP_*}$ is the skew-symmetric matrix that performs the cross product, defined as $\widehat{O_eP_*} v = O_eP_* \times v$.
Therefore, given an external wrench acting on the end-effector $w_{e}$, and assembling the grasp matrices for all the legs as $G = [G_A \enspace G_B \enspace G_C]$, it holds
\begin{equation}
    w_{e} = -G w \enspace.
    \label{eqn:eq_ee_G}
\end{equation}
Because of the constraints imposed by the coupling joints, $w_*~=~H_*^\top \tilde{w_*}$ cannot be arbitrary. The rows of $H_*$ represent a basis for the constraint reaction of the coupling joint. For a coupling revolute joint, since its rotation axis is parallel to $k_3 = \text{T}_b^3(1:3,3)$, we obtain
\begin{equation}
    H_* = 
    \begin{bmatrix}
    \text{T}_b^3(1:3,1:3)^\top & 0_{3 \times 3}\\
    0_{2 \times 3} & \text{T}_b^3(1:3,1:2)^\top\\
    \end{bmatrix} \enspace,
\end{equation}
where the first three columns of $\text{T}_b^{3^\top}$ contain the cartesian components of $\{S_3\}$ versors expressed in $\{S_b\}$. Similarly, the matrix that accounts for the constraints of every leg is
\begin{equation}
    H = 
    \begin{bmatrix}
    H_A & 0_{5\times6} & 0_{5\times6} \\
    0_{5\times6} & H_B & 0_{5\times6} \\
    0_{5\times6} & 0_{5\times6} & H_C
    \end{bmatrix} \enspace .
\end{equation}
Therefore, the static equilibrium of the parallel manipulator holds
\begin{subequations}\label{eqn:static_Preact}
    \begin{empheq}[left=\empheqlbrace]{align}
       & \tau  = J^\top H^\top \tilde{w} \label{eqn:static} \\
       & w_{e} = -GH^\top \tilde{w} \label{eqn:eq_ee} \enspace .  
    \end{empheq}
\end{subequations}  

Since the system is underactuated, it is possible to act directly only on some joints of the manipulator. Therefore, it is convenient to apply a change of base to \eqref{eqn:static} that reorders the joint torques $\tau$, separating the balance on the actuated torques $\tau_a$ from the non-actuated ones $\tau_{\overline{a}}$, as
\begin{equation}
    \begin{bmatrix}
    \tau_a\\
    \tau_{\overline{a}}
    \end{bmatrix}
    =
    \begin{bmatrix}
    B_a\\
    B_{\overline{a}}
    \end{bmatrix} \tau
    = B \tau \enspace,
\end{equation}
where
\begin{equation}
\setcounter{MaxMatrixCols}{12}
    B_a = 
    \begin{bmatrix}
       1 & 0 & 0 & 0 & 0 & 0 & 0 & 0 & 0 & 0 & 0 & 0 \\
       0 & 0 & 0 & 0 & 1 & 0 & 0 & 0 & 0 & 0 & 0 & 0 \\
       0 & 0 & 0 & 0 & 0 & 0 & 0 & 0 & 1 & 0 & 0 & 0
    \end{bmatrix}
\end{equation}
and $B_{\overline{a}}$ completes the orthonormal basis. Finally, left-multiplying \eqref{eqn:static} by the change of base matrix $B$ yields the system of equations (15).

\section*{Appendix B: Derivation of the Elastic Transmission Congruence Equation} \label{appendix: Appendix B}
\begin{figure}[!b]
    \centering
    \includegraphics[width = 0.5\linewidth]{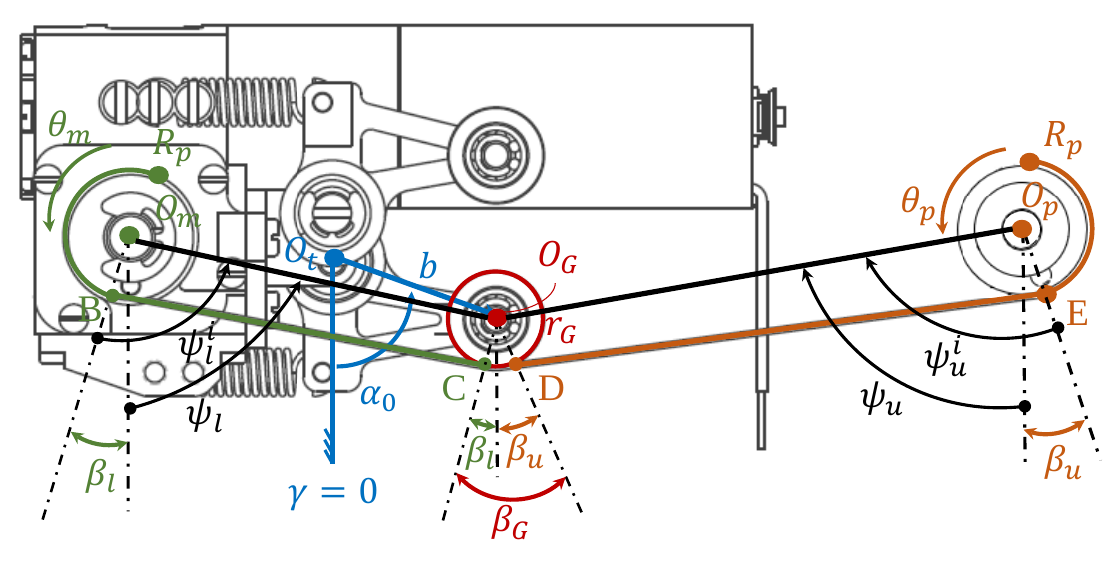}
    \caption{Schematic representation of the elastic transmission mechanism and definition of the geometric parameters required to derive the congruence equation.}
    \label{fig:Appendix_CongruenceTendons}
\end{figure}
This appendix describes the approach employed to derive the model of the elastic transmission mechanism, reported in Section~4.
We derive the function $\delta(\gamma)$ by equating the tendon length $L_t$, known by design, with the same length computed as a function of the geometry of the mechanism (see Figure~\ref{fig:Appendix_CongruenceTendons}), the deflection $\delta$, and the lever angle $\gamma$.

Define the centers of the output motor shaft $O_m$, the fixed lever pulley $O_t$, the first joint of the parallel manipulator $O_P$, and the mobile pulley $O_G$. The winding angle is defined as the angle formed by the normal versor at the contact point of the tendon on the pulley with the fixed horizontal reference.  
While $O_m$, $O_P$, and $O_t$ remain constant and are known by design, the position of $O_G$ depends on $\gamma$ as
\begin{equation}
    O_mO_G(\gamma) = O_mO_t + b 
    \begin{bmatrix}
    cos(\gamma + \alpha_0)\\
    sin(\gamma + \alpha_0)
    \end{bmatrix} \enspace,
\end{equation}
where $b = 18$ mm represents the length of the oblique lever arm and $\alpha_0 = 70.6$° is the interior angle it forms with the horizontal arm. 
By defining the radius of the motor and joint pulleys $R_P$ and that of the mobile pulley as $r_G$, the length of segment BC is
\begin{equation}
    BC(\gamma) = \sqrt{\left|O_mO_G\right|_2^2 - (R_P-r_G)^2} \enspace,
\end{equation}
and the angles $\psi_l$ and $\psi_l^i$ are
\begin{equation}
\hspace{-0.9em}
    \psi_l = acos\left(\frac{(O_mO_G)_x}{\left|O_mO_G\right|_2}\right)  , \enspace
    \psi_l^i = asin\left(\frac{BC}{\left|O_mO_G\right|_2}\right) \,.
\end{equation}
Therefore, the winding angle on the lower side of the mobile pulley $\beta_l$ is their difference $\beta_l = \psi_l^i - \psi_l$.
Similarly, compute the winding angle for the upper side of the mechanism, starting from
\begin{equation}
    O_PO_G(\gamma) = O_PO_t + b 
    \begin{bmatrix}
    cos(\gamma + \alpha_0)\\
    sin(\gamma + \alpha_0)
    \end{bmatrix} \enspace,
\end{equation}
to obtain
\begin{equation}
    DE(\gamma) = \sqrt{\left|O_PO_G\right|_2^2 - (R_P-r_G)^2}
\end{equation}
and the angles $\psi_u$ and $\psi_u^i$ as
\begin{equation}
\hspace{-0.9em}
    \psi_u = acos\left(\frac{(O_PO_G)_x}{\left|O_PO_G\right|_2}\right) , \enspace
    \psi_u^i = asin\left(\frac{DE}{\left|O_PO_G\right|_2}\right) .
\end{equation}

Finally, we compute the winding angle on the upper side of the mobile pulley as $\beta_u = \psi_u^i - \psi_u$.
So, the total winding angle on the mobile lever pulley is $\beta_G = |\beta_l| + \beta_u$, on the motor pulley is $\beta_m = \pi - \theta - |\beta_l|$, and on the joint pulley is $\beta_P = q_1 + \frac{157.5}{180}\pi - \beta_u$.
Now, we can formulate the congruence equation for the right branch of the transmission mechanism by expressing the length of the tendon as a function of the previously computed terms, obtaining
\begin{equation}
    \label{eqn:congruence_cable}
    L_t = R_P \beta_m + BC + r_G\beta_G  + DE +  R_P \beta_P\enspace.
\end{equation}

Substituting $\delta = q_1 - \theta$ into \eqref{eqn:congruence_cable} and solving the equation yield the desired function $\delta(\gamma)$. Next, we compute its derivative $\frac{\partial \delta }{\partial \gamma}$, we invert it, and substitute it into 
\begin{equation}
    \label{eqn:tau_gamma}
   \tau(\gamma) = -\frac{\partial U_{s}}{\partial \delta} = -\frac{\partial U_{s}}{\partial \gamma} \frac{\partial \gamma}{\partial \delta} = -\frac{\partial U_{s}}{\partial \gamma} (\frac{\partial \delta}{\partial \gamma})^{-1} \enspace, 
\end{equation}
which is (20) from the main text, to obtain the torque delivered by the right branch of the elastic transmission on the motor shaft. 
The same procedure can be applied to the left branch of the actuation system. 
Given $\delta$ from the measurements of the encoders and assuming to know $L_t$ by design, it is possible to solve \eqref{eqn:congruence_cable} numerically to obtain $\gamma$, and thus the actuated torque. Finally, accounting for the contribution of the two branches yields the torque delivered to the actuated joint. Since the motor and joint pulleys have the same radii, to achieve the static equilibrium, the torque acting on the motor shaft $\tau_m$ must be of equal magnitude but opposite in verse to the one affecting the actuated joint $\tau_j$, leading to 
\begin{equation}
    \tau_{j} = - \tau_{m} = \tau_{r} + \tau_l \enspace,
\end{equation}
where $\tau_r$ and $\tau_l$ are the torques delivered by the right and left branches, respectively.